\documentclass[journal]{IEEEtran}
\usepackage{amsmath,amsfonts,amsthm}
\usepackage{algorithmic}
\usepackage{algorithm}
\usepackage{array}
\usepackage[colorlinks,urlcolor=blue,linkcolor=blue,citecolor=blue]{hyperref}
\usepackage{textcomp}
\usepackage{stfloats}
\usepackage{url}
\usepackage{verbatim}
\usepackage{graphicx}
\usepackage{cite}
\usepackage{subcaption}
\usepackage[flushleft]{threeparttable}
\usepackage{bbm}
\usepackage[table,dvipsnames]{xcolor}
\usepackage[normalem]{ulem}

\newtheorem{definition}{Definition}

\begin{document}

\title{Incorporating Fairness in Neighborhood Graphs for Fair Spectral Clustering}

\author{Adithya K Moorthy, V. Vijaya Saradhi, and Bhanu Prasad.
\thanks{Adithya K Moorthy and V. Vijaya Saradhi is with Indian Institute of Technology, Guwahati, Assam, India (e-mail: a.k@iitg.ac.in, saradhi@iitg.ac.in).}
\thanks{Bhanu Prasad is with Florida A\&M University, Tallahassee, Florida 32307, USA  (e-mail: bhanu.prasad@famu.edu).}}



\maketitle

\begin{abstract}
    Graph clustering plays a pivotal role in unsupervised learning methods like spectral clustering, yet traditional methods for graph clustering often perpetuate bias through unfair graph constructions that may underrepresent some groups. The current research introduces novel approaches for constructing \emph{fair k-nearest neighbor (kNN)} and \emph{fair $\epsilon$-neighborhood graphs} that proactively enforce demographic parity during graph formation. By incorporating fairness constraints at the earliest stage of neighborhood selection steps, our approaches incorporate proportional representation of sensitive features into the local graph structure while maintaining geometric consistency.
    Our work addresses a critical gap in pre-processing for fair spectral clustering, demonstrating that topological fairness in graph construction is essential for achieving equitable clustering outcomes.
    Widely used graph construction methods like kNN and $\epsilon$-neighborhood graphs propagate edge based disparate impact on sensitive groups, leading to biased clustering results. 
    Providing representation of each sensitive group in the neighborhood of every node leads to fairer spectral clustering results because the topological features of the graph naturally reflect equitable group ratios. 
    This research fills an essential shortcoming in fair unsupervised learning, by illustrating how topological fairness in graph construction inherently facilitates fairer spectral clustering results without the need for changes to the clustering algorithm itself.
    Thorough experiments on three synthetic datasets, seven real-world tabular datasets, and three real-world image datasets prove that our fair graph construction methods surpass the current baselines in graph clustering tasks.
\end{abstract}

\begin{IEEEkeywords}
Fair spectral clustering, k-nearest neighbor graph, $\epsilon$-neighborhood graph, fairness, disparate impact.
\end{IEEEkeywords}

\section{Introduction}

Machine learning algorithms are widely used for decision-making in a variety of fields, including criminal justice \cite{travaini2022machine}, healthcare \cite{henriques2015generative, bandyopadhyay2015data}, and finance \cite{petropoulos2020predicting}.  The reason for this is that these algorithms have been shown to be very accurate and effective at analyzing big datasets.  The increasing prevalence of these algorithms has raised questions regarding their fairness and potential to reinforce societal biases \cite{pmlr-v81-buolamwini18a, o2017weapons}. These biases can result in unfair treatment of certain groups of people thereby create significant societal implications. 

Recently, concerns have been raised about the fairness of clusters produced by popular clustering algorithms. Banks often use datasets, such as \cite{misc_default_of_credit_card_clients_350}, to target customers for loans and financial products. These customers are grouped based on the attributes, including race and gender, in the data. However, systemic biases—such as women earning less salary than men \cite{carnevale2018women} and people of color facing more barriers to education than white individuals \cite{mcgee2020interrogating}—can lead to unequal representation within these clusters. As a result, some groups may receive better loan offers than others. Ensuring that these clusters are equitable is, therefore, crucial. Evidently, the social aspects of clustering are revealed when we consider the data points to be people, and clustering means placing those individuals together based on their attributes. Consider another example, a hiring scenario in which candidates are categorized based on characteristics such as education, experience, and so on. According to these attributes, the candidates are placed into different clusters. In a society where people are segregated as being either part of the black or brown population, a disproportionate representation of one group in a cluster can be recognized as unjust. This case portrays a classic example of disparate impact \cite{feldman2015certifying} in the scenario of clustering. 

Disparate impact is a form of discrimination that occurs when a policy or practice that appears neutral has a disproportionately negative effect on some groups, even if, in that policy or practice, there is no intention to discriminate. In the context of clustering, this can happen when certain groups are overrepresented or underrepresented in the resulting clusters, leading to unfair treatment or outcomes. This concept was brought to the domain of clustering by \cite{Chierichetti_Kumar_Lattanzi_Vassilvitskii_2017} for the purpose of defining equitability in clustering. Since then, various works have adopted this fairness criterion of Balance to enhance impartiality in clustering \cite{Gupta_Dukkipati_2022,liu2022stochastic,bohm2020fair,backurs2019scalable,bercea2018cost, schmidt2020fair,Kleindessner_Samadi_Awasthi_Morgenstern_2019,wang2023scalable,gupta2022consistency}.

Spectral clustering \cite{shi2000normalized} is a widely used clustering algorithm that relies on the construction of a neighborhood graph to capture the relationships between data points. The spectral clustering algorithm uses the eigenvalues and eigenvectors of the graph Laplacian to identify the clusters in the data. 
Recent advances in fair spectral clustering \cite{Kleindessner_Samadi_Awasthi_Morgenstern_2019, wang2023scalable, gupta2022consistency} have effectively addressed algorithmic fairness using in-processing methods that incorporate fairness constraints directly into the clustering algorithms. These methods ensure that the clusters are formed in a way that respects demographic parity, thereby mitigating disparate impact. However, as shown in foundational works \cite{Maier_Luxburg_Hein_2008}, the choice of neighborhood graph construction critically determines spectral clustering outcomes. Thus, this forces us to consider the impact of graph construction on the fairness of clusters obtained from spectral clustering. Critically, this opens up a new avenue for enhancing fairness in spectral clustering by focusing on the graph construction phase.

Traditional kNN and $\epsilon$-neighborhood graphs often create high disparate impact by connecting the nodes preferentially to demographically similar neighbors, reinforcing segregation in the spectral embedding space. This occurs because conventional neighborhood selection purely optimizes geometric proximity, ignoring sensitive attribute distributions. Recent efforts to mitigate bias in graph neural networks \cite{Spinelli_Scardapane_Hussain_Uncini_2022} through edge dropout demonstrate the importance of topological fairness, but such post-processing approaches remain disconnected from core graph construction mechanics. In addition, these methods are not designed to work with spectral clustering, which is one of the most widely used clustering algorithms. 

In the current research, we incorporate fairness directly into popular neighborhood graph construction methods, specifically kNN and $\epsilon$-neighborhood graphs, to ensure that sensitive groups are fairly represented in the local neighborhood structure of each node. We enforce the idea of disparate impact \cite{feldman2015certifying} in the neighborhood of each node, ensuring that the neighborhood of each node has a minimum proportion of nodes from groups different than its own. This approach ensures that the topological structure of the graph reflects equitable group ratios, which is crucial for fair spectral clustering outcomes. 

While constructing the neighborhood graph, we proactively enforce demographic parity by connecting each node to set of neighbors that includes a minimum proportion of nodes from different sensitive groups. This is done by adjusting the neighborhood of each node by adding or removing nodes from the neighborhood until the fairness condition is satisfied. The neighborhood adjustment mechanism ensures adequate representation of each sensitive group in the graph neighborhoods. Fair neighborhoods lead to clustering results where each cluster better reflects the demographic makeup of the overall dataset. This reduces the likelihood that certain clusters are dominated by a single group, which can otherwise result in biased or exclusionary outcomes. By proactively enforcing demographic parity during graph formation, we create a fair kNN graph and a fair $\epsilon$-neighborhood graph that can be used in spectral clustering without requiring post-hoc adjustments or specialized algorithms.  Here, the goal is not to distort outcomes arbitrarily but to create a more equitable starting point that reflects the diversity of the broader population. By enforcing the minimum representation thresholds, the algorithm proposed in the current research prevents the formation of homogenous clusters.

The key contributions of the current research are as follows:
\begin{itemize}
\item Provide a definition to the concept of \textit{fair neighborhoods} in graph construction, grounded on the principle of disparate impact, for ensuring equitable representation of sensitive groups in the local neighborhood structure of each node.
\item Introduce a framework for constructing fair kNN and $\epsilon$-neighborhood graphs that proactively enforce demographic parity during graph formation. This framework ensures that the sensitive groups are adequately represented in the neighborhood of each node, thereby promoting fairness in spectral clustering.
\item Construct the framework as a pre-processing method, which does not require any internal modifications to the spectral clustering algorithm. This makes that framework easy to integrate into existing clustering pipelines.
\item Demonstrate that the fair graph construction methods proposed in the current research outperform the existing fairness-agnostic baselines in spectral clustering tasks, and perform on par or better than the specialized fair clustering algorithms.
\item Demonstrate that the pre-processing methods proposed in the current research performs on par or better than the existing in-processing methods for fair spectral clustering, while being significantly more efficient and easier to implement.
\end{itemize}

The remainder of this research is organized as follows: Section II reviews foundational work in fair clustering and graph construction. Section III explains the notations and preliminaries used in this research, including kNN and $\epsilon$-neighborhood graphs. Section IV introduces the concept of fair neighborhoods. Sections V and VI describe the construction of fair kNN graph and the construction of $\epsilon$-neighborhood graph, respectively. Section VII presents experimental results on synthetic and real-world datasets, demonstrating the effectiveness of our approach. Finally, Section VIII concludes the research.

\section{Related Works}

The general machine learning fairness literature has thoroughly explored disparate impact - a judicially defined phenomenon arising when facially neutral practices have a disproportionate effect on sensitive groups, even in the absence of any discriminatory motive. Initially developed in employment law in landmark cases such as Griggs v. Duke Power Co. (1971), disparate impact has emerged as a standard for measuring algorithmic fairness \cite{spann2009disparate}. For machine learning environments, disparate impact is normally expressed in terms of the 80\% rule, in which the selection rates for unprivileged groups must be no lower than 80\% of the selection rate for privileged groups \cite{feldman2015certifying,barocas2016big}. Recent works have proposed several metrics to measure disparate impact in ML models, such as the Disparate Impact Discrimination Index (DIDI) for regression and classification problems \cite{giuliani2023generalized}, and fairness evaluation frameworks measuring equal opportunity differences and disparate impact ratios between demographic groups \cite{li2023evaluating, shen2022fair}. Research has established that disparate impact can be caused by more than one source in algorithmic systems, such as biased training data, erroneous feature selection, and the combination of different algorithms in various domains \cite{wang2024algorithmic}. This has resulted in the creation of bias mitigation methods like pre-processing data transformations \cite{tawakuli2024make}, in-processing fairness constraints \cite{Kleindessner_Samadi_Awasthi_Morgenstern_2019}, and post-processing calibrations \cite{mishler2021fairness} to counteract disparate impact without compromising the performance of the model.

In the domain of unsupervised learning, Balance is one of the most widely adopted metrics to evaluate fairness. Initially proposed by \cite{Chierichetti_Kumar_Lattanzi_Vassilvitskii_2017}, Balance refers to the distribution of sensitive groups within each cluster. A clustering is considered fair when it maximizes the minimum Balance across all the clusters, thereby ensuring that each cluster mirrors the sensitive group proportions of the entire dataset.

In \cite{Bera_Chakrabarty_Flores_Negahbani_2019}, the authors extend the Balance criteria to scenarios involving multiple sensitive attributes. They introduce two key fairness properties: \textit{Restricted Dominance}, which requires that the proportion of any sensitive group in a cluster does not exceed a predefined threshold, and \textit{Minority Protection}, which ensures that each sensitive group is represented above a minimum threshold in every cluster. These properties are deemed essential for achieving fair clustering under multiple sensitive attributes.

In another line of work, \cite{ahmadian2020fair} adapts the Balance criterion within the agglomerative hierarchical clustering paradigm. That work employs the notion of ``fairlets''—minimal subsets of data points that cannot be further partitioned fairly—as defined in \cite{Chierichetti_Kumar_Lattanzi_Vassilvitskii_2017}. The method in \cite{ahmadian2020fair} aims to optimize both the \textit{value} objective, introduced in \cite{cohen2019hierarchical}, and the \textit{revenue} objective from \cite{NIPS2017_d8d31bd7}. The work in \cite{chhabra2022fair} further contributes to this area by incorporating Balance as a ``fairness cost'' into the hierarchical clustering objective function.

In \cite{zhou2024fair}, the authors propose a novel approach to fair clustering that combines fairness and clustering objectives. They propose a novel approach to fair clustering by introducing a clustering ensemble method that enforces equal cluster capacity, addressing the issue of imbalanced group representation across clusters. Their method ensures fairness by combining multiple clustering results while maintaining strict capacity constraints, thereby enhancing both fairness and clustering quality. This contribution is particularly relevant in scenarios where equitable treatment of subgroups is essential, such as in social data analysis or resource allocation tasks.

Research on fairness in graph-based clustering is still emerging. The work in \cite{Kleindessner_Samadi_Awasthi_Morgenstern_2019} brings the Balance constraint into the spectral clustering framework. Another innovative approach is proposed by \cite{gupta2022consistency}, which combines both group and individual fairness using a representation graph. That research introduces a representation-aware variant of spectral clustering and provides theoretical consistency guarantees. The fair spectral clustering algorithm was extended to dynamic graphs in \cite{fu2023fairness}.

A key limitation in the current fair spectral clustering techniques is that, firstly, the in-processing methods are mostly expensive to design and deploy in practice. Secondly, they often overlook the critical role of graph construction in shaping clustering outcomes. The choice of graph construction method, such as kNN or $\epsilon$-neighborhood graphs, significantly impact the spectral embedding and, consequently, the clustering results \cite{Maier_Luxburg_Hein_2008}.

\section{Preliminaries}\label{sec:preliminaries}

Some notations used in this research are provided in Table~\ref{table:notations}. Bold uppercase letters (e.g., \textbf{Q}) denote matrices and bold lowercase letters (e.g., \textbf{q}) denote vectors. Lowercase italic letters (e.g., $q$) stand for scalar values.

\begin{table*}[t]
    \caption{Notations}
    \centering
    \label{table:notations}
    \begin{threeparttable}
    \begin{tabular}{ll}
      \cline{1-2}
      \textbf{Symbol}               & \textbf{Definition}                                                                                                  \\
      \cline{1-2}
      $d$                           & Number of features in the dataset                                                                                   \\
      $\mathcal{X}$                  & $\{\textbf{x}_1, \textbf{x}_2, ..., \textbf{x}_n\}$; $\textbf{x}_i \in \mathbb{R}^{d}$  \\
      $V$                           & Set of vertices in the graph given by $V = \{v_1, v_2, \ldots, v_n\}$, where $v_i$ corresponds to $\textbf{x}_i$ in the dataset $\mathcal{X}$ \\
      $E$                           & Set of edges in the graph, where $(v_i, v_j) \in E$ indicates an edge between vertices $v_i$ and $v_j$ \\
      $G(V,E)$                      & Graph $G$ with vertex set $V$ and edge set $E$ \\
      $n , m$                       & $n$ is the number of vertices in $G$ and $m$ is the number of edges in $G$                                                 \\
      $\textbf{W}_{n \times n}$     & Weighted similarity matrix with $w_{ij}$ as weight of an edge between $\textbf{x}_{i}$ and $\textbf{x}_{j}$ \\
      $c$                           & Number of clusters                                                                                                   \\
      $k$                           & Number of nearest neighbors in a kNN algorithm                                                       \\
      $\varepsilon$                  & Radius of the neighborhood in an $\varepsilon$-neighborhood graph                                                     \\
        $\mathcal{N}_k(\textbf{x}_i)$ & Set of $k$ nearest neighbors of $\textbf{x}_i$ in a kNN graph                                                        \\
        $\mathcal{N}_\varepsilon(\textbf{x}_i)$ & Set of neighbors of $\textbf{x}_i$ in an $\varepsilon$-neighborhood graph                                            \\
      $\sigma$                      & Scaling parameter for Gaussian kernel in edge weighting                                                       \\
      $h$ & Number of values the sensitive attribute can take or the number of sensitive groups \\
      $\alpha$ & Fairness parameter controlling the minimum cross-group representation ratio in fair neighborhoods \\
      \cline{1-2}
    \end{tabular}
    \begin{tablenotes}
      \small
      \item $\{x_i, x_j\}$ and $\{v_i, v_j\}$ are used interchangeably to denote the same data point in the dataset and the graph, respectively.
    \end{tablenotes}
  \end{threeparttable}
  \end{table*}

\subsection{kNN Graph}

The construction of similarity graphs forms a foundational step in numerous machine learning and data analysis tasks, particularly in unsupervised learning. Among various types of graphs, the \emph{$k$-nearest neighbor (kNN) graph} is widely utilized due to its computational efficiency and ability to capture local structure in high-dimensional datasets. 
In this section, we introduce the basic concepts and notation associated with kNN graph construction, and outline its significance as a precursor to advanced graph-based algorithms.

For a dataset $\mathcal{X}$, kNN graph is constructed by associating each point $\textbf{x}_i$ with edges to its $k$ closest neighbors in $\mathcal{X}$ under a given distance metric, typically the Euclidean distance. The result is a sparse graph $G = (V, E)$, where $(\textbf{x}_i, \textbf{x}_j) \in E$ if $\textbf{x}_j$ is among the $k$ nearest neighbors of $\textbf{x}_i$.

Several variants of kNN graphs exist, depending on the symmetry and weighting of edges. In the \emph{directed} kNN graph, an edge from $\textbf{x}_i$ to $\textbf{x}_j$ is included if $\textbf{x}_j \in \mathcal{N}_k(\textbf{x}_i)$, where $\mathcal{N}_k(\textbf{x}_i)$ denotes the $k$ nearest neighbors of $\textbf{x}_i$. The \emph{mutual} or \emph{symmetric} kNN graph includes an edge $(\textbf{x}_i, \textbf{x}_j)$, only if both $\textbf{x}_i \in \mathcal{N}_k(\textbf{x}_j)$ and $\textbf{x}_j \in \mathcal{N}_k(\textbf{x}_i)$. A \emph{fully symmetrized} version adds an edge if either of the two nodes considers the other a neighbor. Edge weights may also be introduced, commonly using a Gaussian kernel:
\begin{equation}
w_{ij} = \exp\left(-\frac{\|\textbf{x}_i - \textbf{x}_j\|^2}{2\sigma^2}\right)
\end{equation}
where $\sigma$ is a scaling parameter that controls the influence of local distances.

The kNN graph is particularly useful in spectral clustering, where the graph Laplacian is computed to identify the clusters based on eigenvector analysis. The sparsity of the kNN graph ensures computational efficiency while preserving local structure, making it suitable for large datasets.

\subsection{$\epsilon$-Neighborhood Graph}

The \emph{$\epsilon$-neighborhood graph} provides an alternative paradigm for constructing the similarity graphs by leveraging absolute distance thresholds rather than relative neighbor counts. This approach is particularly effective in scenarios where uniform density assumptions hold or when explicit control over connection radii is desired.

The $\epsilon$-neighborhood graph is defined as the graph $G = (V, E)$ and $(\textbf{x}_i, \textbf{x}_j) \in E$ if $\|\textbf{x}_i - \textbf{x}_j\| \leq \epsilon$, where $\|\cdot\|$ denotes a distance metric (commonly Euclidean). This construction results in a graph where edges are formed based on proximity within a fixed radius $\epsilon$, allowing for flexible neighborhood sizes that adapt to local data density.
This construction induces symmetric edges when using standard metrics like Euclidean distance, though directed variants may emerge with asymmetric distance measures.

Even in an $\epsilon$-neighborhood graph, edge weighting typically follows Gaussian kernel to emphasize local relationships:
\begin{equation}
w_{ij} = \begin{cases} 
\exp\left(-\frac{\|\textbf{x}_i - \textbf{x}_j\|^2}{2\sigma^2}\right) & \text{if } \|\textbf{x}_i - \textbf{x}_j\| < \epsilon \\
0 & \text{otherwise}
\end{cases}
\end{equation}
The resulting graph adjacency matrix exhibits block-diagonal dominance under cluster-separated distributions, making it particularly suitable for spectral methods.

\begin{figure*}
\centering
\begin{subfigure}[b]{0.5\textwidth}
    \centering
    \includegraphics[width=0.35\textwidth]{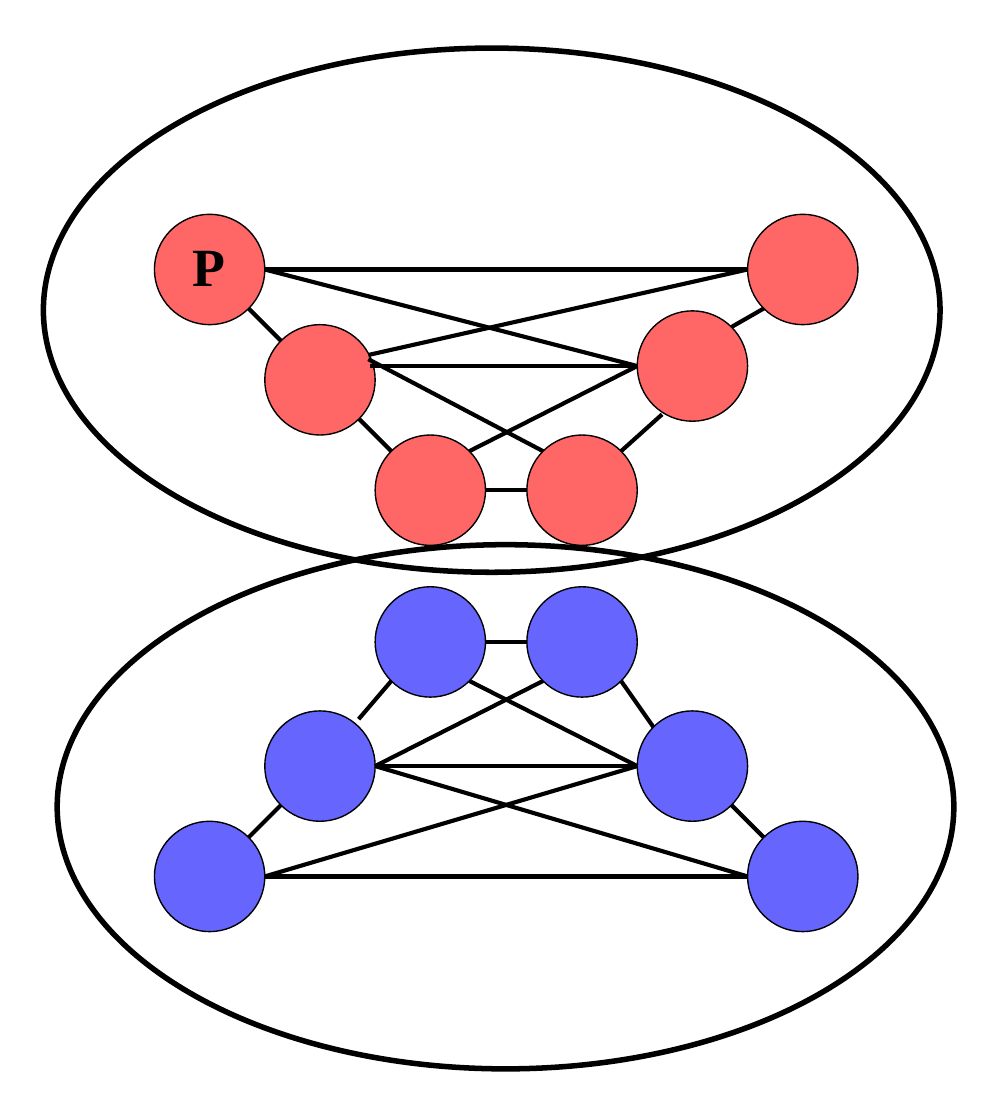}
    \hspace{0.5cm}
    \includegraphics[width=0.35\textwidth]{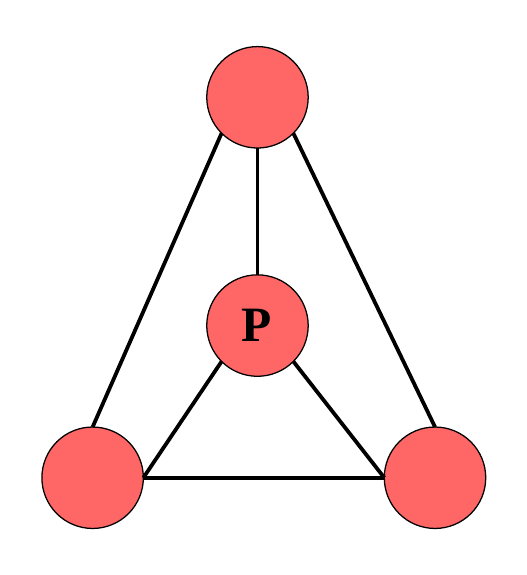}
    \caption{}
\end{subfigure}
\begin{subfigure}[b]{0.5\textwidth}
    \centering
    \includegraphics[width=0.35\textwidth]{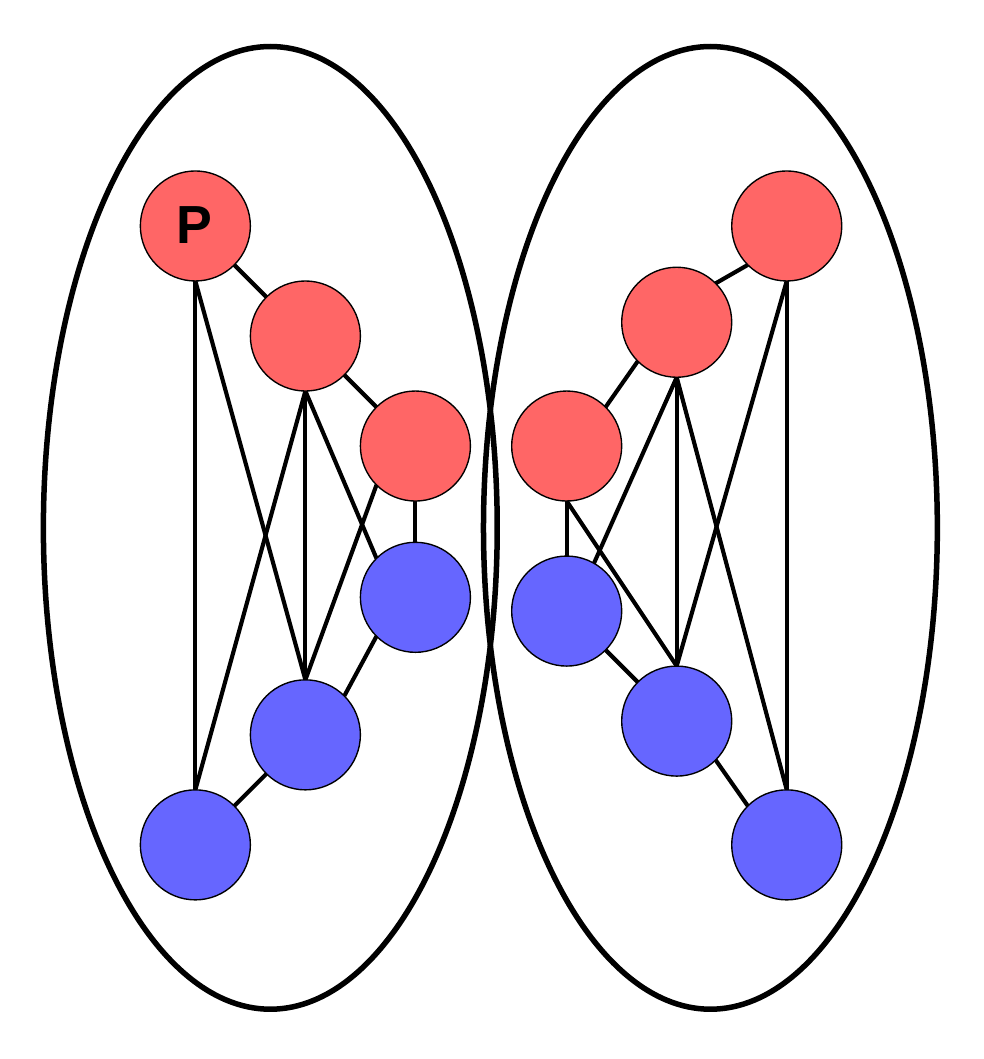}
    \hspace{0.5cm}
    \includegraphics[width=0.35\textwidth]{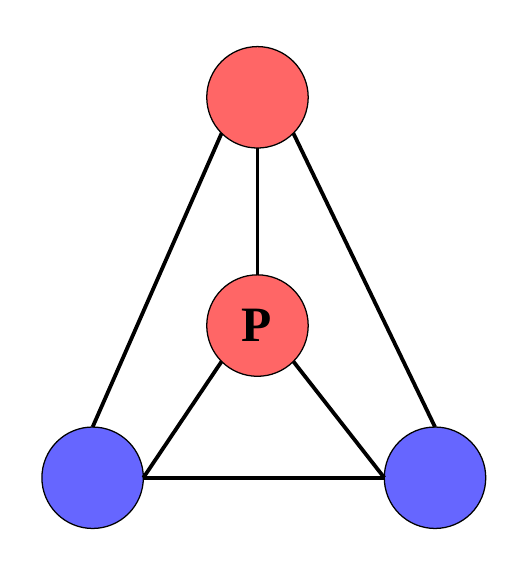}
    \caption{}
\end{subfigure}
\caption{Left side of Figure (a) shows a graph constructed with unfair neighborhoods, resulting in a biased clustering outcome. A portion of the unfair neighborhood is shown on the right side. Left side of Figure (b) shows a graph constructed with fair neighborhood, leading to a more balanced clustering outcome. A portion of that fair neighborhood is shown on the right side.} 
\label{fig:fair_unfair_neighborhood}
\end{figure*}

\subsection{Density Reachability}  
Density reachability constitutes a fundamental concept in density-based clustering algorithms, particularly in DBSCAN (Density-Based Spatial Clustering of Applications with Noise)\cite{dbscan}, where it governs the expansion and connectivity of clusters. This principle formalizes how the points within a dataset propagate the cluster membership through localized density conditions, enabling the discovery of arbitrarily shaped structures while robustly handling the noise.

\begin{definition}[Core Point]

A point $\textbf{x}_i \in \mathcal{X}$ is designated as a core point if its $\varepsilon$-neighborhood contains at least $min\_pts$ number of points, including itself:
$$
|\mathcal{N}_\varepsilon(\textbf{x}_i)| \geq min\_pts,
$$
where $\mathcal{N}_\varepsilon(\textbf{x}_i) = \{\textbf{x}_j \in \mathcal{X} \mid \|\textbf{x}_i - \textbf{x}_j\| \leq \varepsilon\ \forall j = \{1 \dots n\} \}$.

\end{definition}

From this foundation, we define two key relationships that underpin density reachability:
\begin{definition}[Direct Density Reachability]
  Direct density reachability: A point $\textbf{x}_j$ is directly density-reachable from $\textbf{x}_i$ if
\begin{enumerate}
   \item $\textbf{x}_j \in \mathcal{N}_\varepsilon(\textbf{x}_i)$  
   \item $\textbf{x}_i$ is a core point.  
\end{enumerate}  
\end{definition}

\begin{definition}[Density reachability]
   A point $\textbf{x}_q$ is density-reachable from $\textbf{x}_p$ if there exists a chain $\{\textbf{x}_p = \textbf{x}_1, \textbf{x}_2, \ldots, \textbf{x}_m = \textbf{x}_q\}$ where each $\textbf{x}_{i+1}$ is directly density-reachable from $\textbf{x}_i$. 
\end{definition}

These definitions establish the foundation for density-based clustering, enabling the identification of clusters as maximal sets of mutually density-reachable points. Here, we use the concept of density reachability to ensure that the nodes added to the neighborhood of another node, to facilitate fairness do not tarnish the local density of that neighborhood.

\section{Fair Neighborhood Construction}

In the previous section, we discussed purely geometric definitions of kNN and $\varepsilon$-neighborhood graphs, which do not consider sensitive attributes. In this section, we introduce fairness into the neighborhood graph construction process, specifically focusing on kNN and $\varepsilon$-neighborhood graphs.
We introduce a novel approach for constructing \textbf{fair neighborhood graphs}, where fairness with respect to sensitive group representation is enforced within the local neighborhood structure of each node. This method aims to mitigate bias in downstream tasks such as spectral clustering, where standard neighborhood graphs may inadequately represent sensitive groups, leading to unfair outcomes. By ensuring an adequate representation of sensitive groups in the neighborhood of each node, our approach promotes fairness throughout the clustering process. An illustration of the unfair and fair neighborhoods is shown in Figure~\ref{fig:fair_unfair_neighborhood}.

\begin{figure*}
\centering
    \includegraphics[width=\textwidth]{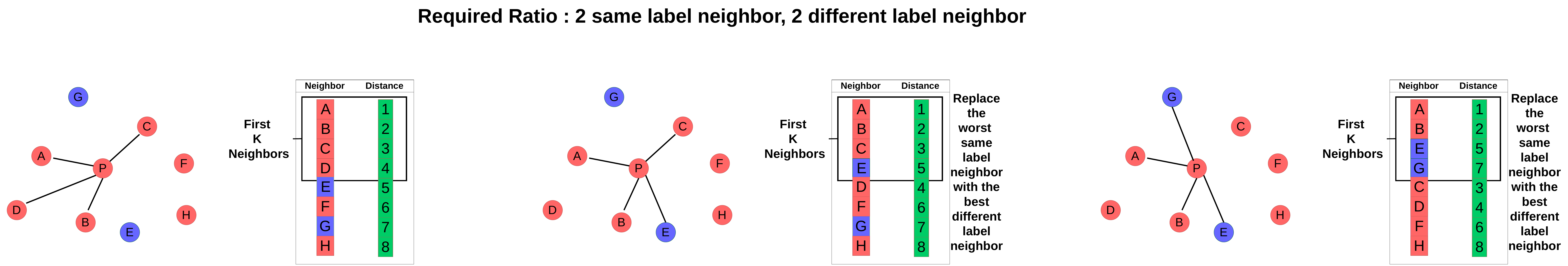}
\caption{From the left, assuming $k=4$, the first figure shows the initial kNN graph constructed with unfair neighborhoods, leading to a biased clustering outcome. The second figure illustrates the process of adjusting neighborhoods to ensure fairness, where nodes from the same sensitive group are replaced with nodes from a different sensitive group. The third figure shows the final fair kNN graph, which leads to a more balanced neighborhood.} 
\label{fig:fair_knn_steps}
\end{figure*}

For a graph $G$, each node \( v_i \in V \) is associated with a sensitive attribute \( s(v_i) \), indicating its membership in one of \( h \) predefined sensitive groups \( S_1, S_2, \dots, S_h \). For example, consider an attribute color, which can take values red, green, or blue. The sensitive attribute \( s(v_i) \) for a node \( v_i \) could be defined as the color of the node, such that \( s(v_i) \in \{ \text{red}, \text{green}, \text{blue} \} \).
This sensitive attribute is leveraged to enforce fairness during graph construction.

We define the \textit{neighborhood} of a node \( v_i \), denoted \( \mathcal{N}_{\ast}(v_i) \), as a subset of \( V \setminus \{v_i\} \) selected based on a similarity measure. Here, \( \ast \) can represent either a $k$, indicating a kNN graph, or \( \varepsilon \), indicating an $\varepsilon$-neighborhood graph.  
The construction of \( \mathcal{N}_{\ast}(v_i) \) may follow either:

\begin{itemize}
    \item a \textbf{k-nearest neighbor} rule, where the top \( k \) most similar nodes are selected, or
    \item an \textbf{\( \varepsilon \)-neighborhood} rule, where all nodes within a similarity threshold \( \varepsilon \) are included.
\end{itemize}

Formally,
\begin{multline}
\mathcal{N}_{\ast}(v_i) = 
\begin{cases}
\text{top-}k\left\{v_j \in V \setminus \{v_i\} \text{ ordered by } \| v_i - v_j\| \right\} \\
\quad \text{(kNN)} \\
\left\{v_j \in V \setminus \{v_i\} \mid \| v_i - v_j \| \leq \varepsilon\right\} \\
\quad \text{($\varepsilon$-neighborhood)}
\end{cases}
\end{multline}
where, \( \text{top-}k \) selects the \( k \) nodes with the highest similarity scores.

We impose mitigating disparate impact as a \textbf{fairness condition} on \( \mathcal{N}_{\ast}(v_i) \) to ensure that sensitive groups are adequately represented. Specifically, for each node \( v_i \), the neighborhood should contain a minimum proportion of nodes from groups different than its own. 

\begin{definition}[Fair Neighborhood] \label{def:fair_neighborhood}
  Let \( \alpha \in [0, 1] \) be a user-defined parameter controlling the required cross-group representation. A neighborhood \( \mathcal{N}_{\ast}(v_i) \) of node \( v_i \) is considered \textbf{fair} if it satisfies the following condition:

\begin{multline}\label{eq:fairness_condition}
\frac{|\{ v_j \in \mathcal{N}_{\ast}(v_i) \mid s(v_j) \neq s(v_i) \}|}{|\mathcal{N}_{\ast}(v_i)|} \\
\geq \alpha \cdot \frac{|\{ v_j \in \mathcal{N}_{\ast}(v_i) \mid s(v_j) = s(v_i) \}|}{|\mathcal{N}_{\ast}(v_i)|}
\end{multline}
\end{definition}

If this condition is not satisfied for a given node, the neighborhood \( \mathcal{N}_{\ast}(v_i) \) is adjusted—typically by replacing some same-group neighbors with nodes from other groups—until the fairness criterion is met.

This generalized framework applies to kNN graph construction or $\epsilon$-neighborhood graph construction strategy and provides a tunable mechanism for integrating fairness into graph-based learning pipelines. With the idea of fair neighborhoods in place, we can now proceed to construct fair kNN and fair \( \varepsilon \)-neighborhood graphs. The following sections detail how nodes are added or removed from the neighborhood to ensure fairness.

\section{Introducing fair neighborhoods to kNN graphs}

Building upon the foundation of fair neighborhood construction, we present a unified framework for developing fair k-nearest neighbor (fair kNN) graphs that intrinsically enhances fairness in spectral clustering tasks. 
The core innovation lies in dynamically adjusting neighborhood composition during graph construction to enforce proportional representation of sensitive groups, thereby creating an equitable topological structure for downstream clustering tasks. Thus, we define a fair kNN graph as a kNN graph where each node's neighborhood satisfies the fairness condition defined in Equation~\ref{eq:fairness_condition}. 

To construct a fair adjacency matrix through constrained neighbor selection that simultaneously optimizes for geometric proximity and demographic parity, for each node $ \mathbf{x}_i $, define its candidate neighborhood $ \mathcal{C}_{k}(v_i) $ as the union of its $ k' $-nearest neighbors ($ k' > k $) from all sensitive groups:  

\begin{equation}
\mathcal{C}_{k'}(v_i) = \bigcup_{l=1}^h \left\{ \mathbf{x}_j \in \mathbf{X}_l \mid \text{rank}(\|\mathbf{x}_i - \mathbf{x}_j\|) \leq k' \right\}
\end{equation}
where, $ \mathbf{X}_l $ is the set of nodes in sensitive group $ l \in \{1, \dots, h\} $, and $ \text{rank} (\cdot) $ orders neighbors by ascending distance. The candidate neighborhood $ \mathcal{C}_{k'}(v_i) $ contains the top $ k' $ nearest neighbors across all groups, ensuring a diverse pool for selection. The final neighborhood $ \mathcal{N}_{k}(v_i) \subset \mathcal{C}_{k'}(v_i) $ of size $ k $ is selected to satisfy Balance constraints while minimizing local distortion. Let us define the same-group and different-group sets for each node as \( \mathcal{S}(v_i) = \{ v_j \in V \setminus \{v_i\} \mid s(v_j) = s(v_i) \} \) and \( \mathcal{D}(v_i) = \{ v_j \in V \setminus \{v_i\} \mid s(v_j) \neq s(v_i) \} \), respectively. For the nodes in the same sensitive group \( \mathcal{S}(v_i) \) and different sensitive group \( \mathcal{D}(v_i) \), the composition adheres to:

\begin{equation}
\frac{|\mathcal{N}_{k}(v_i) \cap \mathcal{D}(v_i)|}{|\mathcal{N}_{k}(v_i) \cap \mathcal{S}(v_i)|} \geq \alpha \quad \forall \; v_i \in V
\end{equation}
with $ \alpha \in [0,1] $ controlling the minimum cross-group representation ratio. This is a modified representation of the fairness condition defined in Equation~\ref{eq:fairness_condition}. 
The selection process is solving a constrained optimization like:  

\begin{equation}
\begin{aligned}
\mathcal{N}_{k}(v_i)^* = \arg\min_{\mathcal{N}_{k}(v_i) \subseteq \mathcal{X} \setminus \{\mathbf{x}_i\}} \ & \sum_{\mathbf{x}_j \in \mathcal{N}_{k}(v_i)} \|\mathbf{x}_i - \mathbf{x}_j\|^2 \\
\text{s.t.} \quad & |\mathcal{N}_{k}(v_i)| = k \\
& \frac{|\mathcal{N}_{k}(v_i) \cap \mathcal{D}(v_i)|}{|\mathcal{N}_{k}(v_i) \cap \mathcal{S}(v_i)|} \geq \alpha \\
\end{aligned}
\label{eq:constrained-knn}
\end{equation}

\begin{algorithm}[t]
\caption{Fair NN-Descent}
\label{alg:fair_nn_descent}
\begin{algorithmic}[1]
\REQUIRE Dataset $\mathbf{X}$, number of neighbors $k$, group labels $s$, fairness ratio $\alpha$
\ENSURE Neighbor lists $\mathcal{N}[v] \; \forall \; v \in V$
\STATE $\mathcal{N}[v] \gets \textsc{Sample}(V, k) \times \{\infty\}, \; \forall \; v \in V$
\LOOP
    \STATE $\mathcal{R} \gets \textsc{Reverse}(\mathcal{N})$
    \STATE $\mathcal{N}[v] \gets \mathcal{N}[v] \cup \mathcal{R}[v], \; \forall \; v \in V$
    \STATE $c \gets 0$
    \FOR{$v \in V$}
        \FOR{$u_1 \in \mathcal{N}[v],\ u_2 \in \mathcal{N}[u_1]$}
            \STATE $\ell \gets \| v - u_2\|$
            \STATE $c \gets c + \textsc{Fair\_update}(\mathcal{N}[v], \langle u_2, \ell, s[u_2] \rangle, \alpha)$
        \ENDFOR
    \ENDFOR
    \IF{$c = 0$}
        \STATE \textbf{return} $\mathcal{N}$
    \ENDIF
\ENDLOOP
\end{algorithmic}
\end{algorithm}

This process is implemented through an iterative replacement mechanism. Let $ \mathcal{N}_{k}(v_i)^{(t)} $ denotes the neighborhood at iteration $ t $, initialized as the standard $ k $-NN set. For each $ \mathbf{x}_i \in \mathbf{X} $, while the Balance ratio violates $ \alpha $, we perform the following steps to update the neighborhood:  

\begin{enumerate}
\item Identify the same-group neighbor with the largest distance:  
   $$
   \mathbf{s}^* = \arg\max_{\mathbf{x}_j \in \mathcal{N}_{k}(v_i)^{(t)} \cap \mathcal{S}(v_i)} \|\mathbf{x}_i - \mathbf{x}_j\|
   $$  

\item Identify different-group candidate with the smallest distance:  
   $$
   \mathbf{d}^* = \arg\min_{\mathbf{x}_j \in (\mathcal{C}_i \setminus \mathcal{N}_{k}(v_i)^{(t)}) \cap \mathcal{D}(v_i)} \|\mathbf{x}_i - \mathbf{x}_j\|
   $$  

\item Update $ \mathcal{N}_{k}(v_i)^{(t+1)} = (\mathcal{N}_{k}(v_i)^{(t)} \setminus \{\mathbf{s}^*\}) \cup \{\mathbf{d}^*\}$ . 

\end{enumerate}

The process terminates when all neighborhoods satisfy Equation~\ref{eq:fairness_condition}.

\begin{algorithm}[t]
\caption{FAIR\_UPDATE}
\label{alg:fair_update}
\begin{algorithmic}[1]
\REQUIRE Current list $\mathcal{N}[v]$, candidate $\langle u, \ell, s[u] \rangle$, $\alpha$
\ENSURE Updated $\mathcal{N}[v]$ if fairness is satisfied
\STATE Count same-group and different-group neighbors
\IF{$|\mathcal{N}[v]| < k$}
    \STATE Add $u$ to $\mathcal{N}[v]$; \textbf{return} $1$
\ENDIF
\IF{$s[u] = s[v]$}
    \STATE Add if max same-group not exceeded. If exceeded, replace the worst same-group neighbor with $u$.
\ELSE
    \STATE if ratio is satisfied, replace worst neighbor. Else, replace the worst same-group neighbor with $u$.
\ENDIF
\STATE \textbf{return} $1$ if updated, else $0$
\end{algorithmic}
\end{algorithm}

An illustration of this process is shown in Figure~\ref{fig:fair_knn_steps}, where the left side shows the unfair kNN graph, the middle shows the adjustment process, and the right side shows the final fair kNN graph.

The Balance-constrained neighborhood construction, proposed in the current research, is seamlessly incorporated into the kNN graph via a modified NN-Descent algorithm \cite{dong2011efficient}. In this implementation, the standard neighbor update mechanism of NN-Descent is augmented to enforce the proportional representation constraint during each iteration of neighborhood refinement. Specifically, when candidate neighbors are evaluated for inclusion, the algorithm dynamically maintains the required ratio between similar group and dissimilar group members in each node's neighborhood, as dictated by the value of parameter $\alpha$. This is achieved by tracking the group composition of current neighbors and selectively replacing or retaining candidates to ensure that Equation~\ref{eq:fairness_condition} is satisfied alongside geometric proximity. The algorithm is outlined in Algorithm~\ref{alg:fair_nn_descent}. The \textsc{Fair\_update} function used in Algorithm~\ref{alg:fair_nn_descent} is detailed in Algorithm~\ref{alg:fair_update}.

By incorporating the fairness-aware selection directly into the iterative neighbor discovery and update steps, the resulting kNN graph simultaneously optimizes for both local similarity and demographic Balance. This approach leverages the efficiency and scalability of NN-Descent, allowing the construction of large-scale fair kNN graphs without post-processing or significant computational overhead. As a result, the topological fairness guarantees are preserved throughout the graph construction process and naturally propagate to downstream spectral clustering tasks.

\section{Introducing Fair Neighborhoods to $\epsilon$-Neighborhood Graphs}

\begin{figure*}
\centering
\begin{subfigure}[b]{0.32\textwidth}
    \centering
    \includegraphics[width=.7\textwidth]{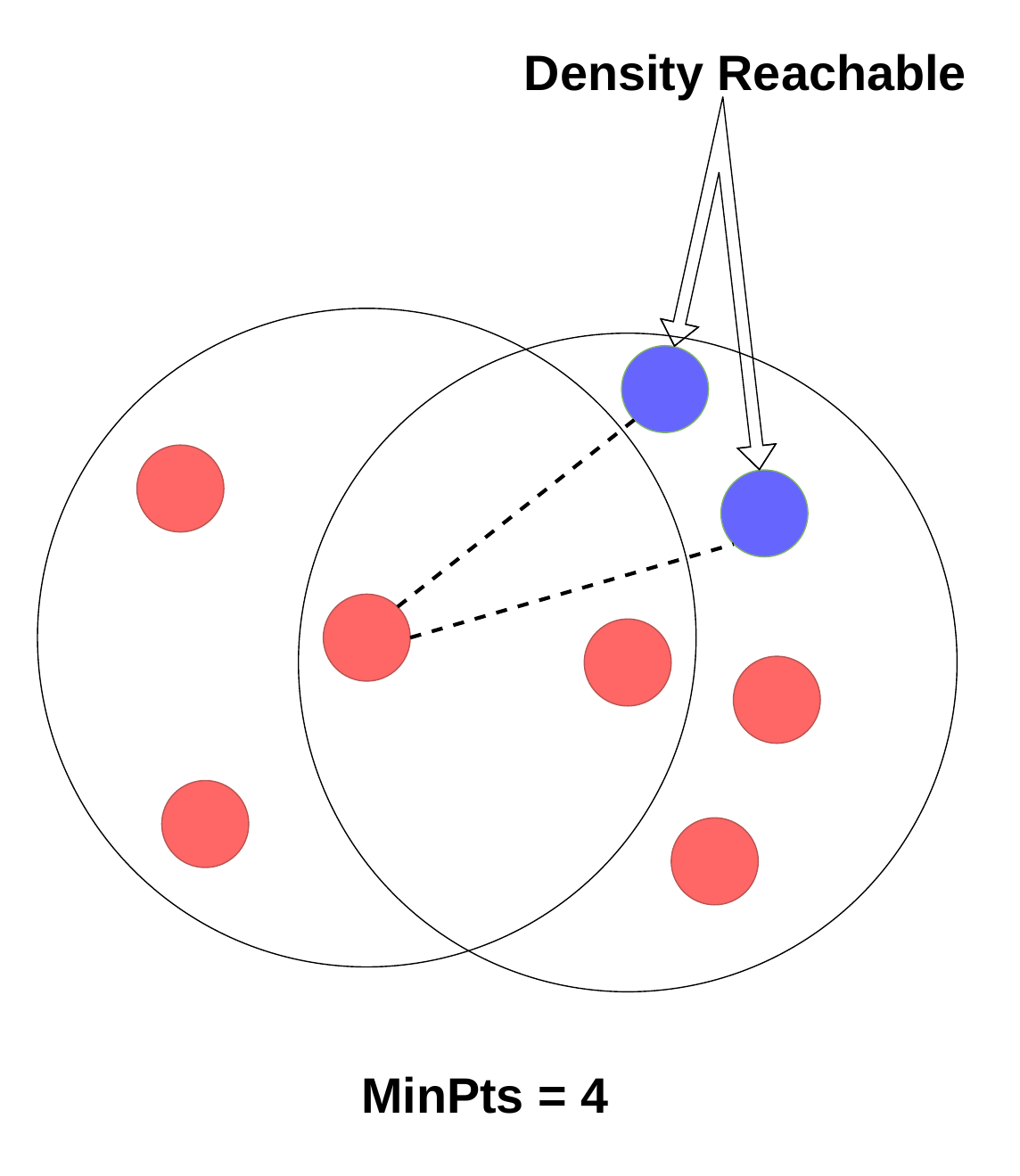}
\end{subfigure}
\begin{subfigure}[b]{0.32\textwidth}
    \centering
    \includegraphics[width=0.6\textwidth]{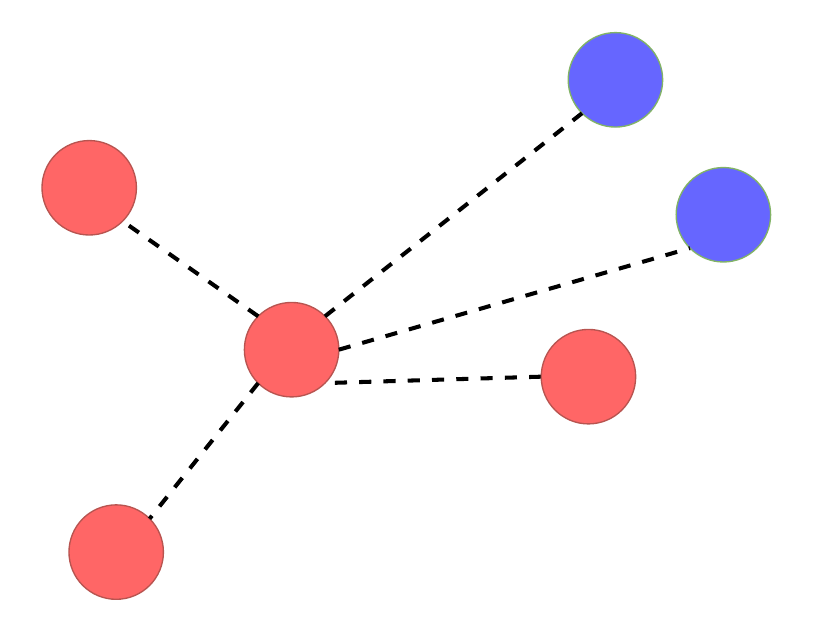}
    \vspace{0.9cm}
\end{subfigure}

\caption{The process of constructing fair $\epsilon$-neighborhood graphs. The left side shows the unfair neighborhood graph and how it is adjusted to ensure fairness. The right side shows the fair neighborhood graph after adjustments. The nodes are colored based on their sensitive group membership, and the dotted lines represent edges in the neighborhood graph.} 
\label{fig:fair_epsilon_steps}
\end{figure*}

Building on the principles established for fair kNN graphs, we extend the fair neighborhood construction paradigm to $\epsilon$-neighborhood graphs through density-aware adjustments that enforce proportional representation of sensitive groups. This approach leverages the inherent connectivity of density-based spatial clustering to maintain geometric coherence while satisfying Equation~\ref{eq:fairness_condition}.

The $\epsilon$-neighborhood graph $G = (V, E)$ is initially constructed through radial proximity:

\begin{equation}
\mathcal{N}_{\epsilon}(v_i) = \{ v_j \in V \setminus \{v_i\} \mid \|\mathbf{x}_i - \mathbf{x}_j\| \leq \epsilon \}
\end{equation}

Similar to the kNN graph, each node $v_i$ is associated with a sensitive attribute $s(v_i)$, indicating its membership in one of $h$ sensitive groups. The goal is to ensure that the neighborhood $\mathcal{N}_{\epsilon}(v_i)$ contains a fair representation of nodes from different sensitive groups.
To achieve this, we define a fairness ratio $\alpha \in [0, 1]$ that controls the minimum proportion of nodes from different sensitive groups in each neighborhood. The fairness condition is defined in Definition~\ref{def:fair_neighborhood}.
If the neighborhood $\mathcal{N}_{\epsilon}(v_i)$ does not satisfy Equation~\ref{eq:fairness_condition}, we adjust it by replacing some same-group neighbors with nodes from different groups. The adjustment process is similar to that of kNN graphs, but it operates on the density-connected structure of the $\epsilon$-neighborhood graph.
Exactly, for each node $v_i$, we perform the following steps:
\begin{itemize}
    \item Count the number of same-group and different-group neighbors in $\mathcal{N}_{\epsilon}(v_i)$.
    \item If the fairness condition is not satisfied, identify the set of reachable nodes $R$ from $v_i$ using density reachability.
    \item Add up to $r$ new nodes with a different sensitive group from $R$, prioritized by smallest hop lengths, to $\mathcal{N}_{\epsilon}(v_i)$, where $r$ is determined by the difference between the required and current number of different-group neighbors.
    \item Repeat until the fairness condition is satisfied.
\end{itemize}
An overview of this process is illustrated in Figure~\ref{fig:fair_epsilon_steps}, where the left side shows the unfair neighborhood graph and the right side shows the fair neighborhood graph after adjustments.
This process ensures that the neighborhood of each node is adjusted to maintain fairness while preserving the density-based connectivity of the graph. The algorithm for constructing fair $\epsilon$-neighborhood graphs is outlined in Algorithm~\ref{alg:fair_epsilon_graph}.

\begin{algorithm}[t]
\caption{Fair $\varepsilon$-Neighborhood Graph Construction}
\label{alg:fair_epsilon_graph}
\begin{algorithmic}[1]
\REQUIRE Dataset $\mathcal{X} = \{\mathbf{x}_1, \ldots, \mathbf{x}_n\}$, sensitive group labels $s$, radius $\varepsilon$, minimum points $minPts$, fairness parameter $\alpha$
\ENSURE Fair $\varepsilon$-neighborhood graph $G(V, E)$
\STATE Initialize $V = \{v_1, \ldots, v_n\}$, where $v_i$ corresponds to $\mathbf{x}_i \in \mathcal{X}$
\STATE Initialize edge set $E = (v_i, v_j)$ for all pairs $(i, j)$ such that $\|\mathbf{x}_i - \mathbf{x}_j\| \leq \varepsilon$
\FOR{each vertex $v_i \in V$}
    \STATE Compute $\mathcal{N}_\varepsilon(\mathbf{x}_i) = \{\mathbf{x}_j \in \mathcal{X} \setminus \{\mathbf{x}_i\} \mid \text{dist}(\mathbf{x}_i, \mathbf{x}_j) \leq \varepsilon\}$
    \STATE Count same-group and different-group neighbors in $\mathcal{N}_\varepsilon(\mathbf{x}_i)$
    \IF{Equation~\ref{eq:fairness_condition} is not satisfied}
        \STATE Let $r$ be the number of additional different-group neighbors needed to satisfy Equation~\ref{eq:fairness_condition}
        \STATE Let $R$ be the set of density-reachable points from $\mathbf{x}_i$
        \STATE Add $r$ closest different-group nodes from $R$ to $\mathcal{N}_\varepsilon(\mathbf{x}_i)$
    \ENDIF
    \FOR{each $\mathbf{x}_j \in \mathcal{N}_\varepsilon(\mathbf{x}_i)$}
        \STATE Add edge $(v_i, v_j)$ to $E$
    \ENDFOR
\ENDFOR
\RETURN $G(V, E)$
\end{algorithmic}
\end{algorithm}

The hop distance serves as a tiebreaker, prioritizing nodes that are closer in the density-connected structure. This ensures minimal distortion of the original spatial relationships while satisfying Equation~\ref{eq:fairness_condition}.

The augmented neighborhoods induce an adjacency matrix where connections represent both geometric proximity and demographic proportionality. When employed in spectral clustering, this graph structure propagates fairness constraints through the graph Laplacian $L = D - A$, ensuring that the resulting spectral embedding preserves both cluster coherence and group Balance.

The unified treatment of kNN and $\epsilon$-neighborhood graphs through their respective fairness mechanisms demonstrates the flexibility of topological constraint propagation in graph-based clustering. Both approaches share the fundamental principle of modifying local neighborhood structures to encode global fairness objectives, which then naturally influence the spectral embedding space through the graph's Laplacian matrix.

\section{Results and Discussion}

In this section, we report the results of our experiments conducted on both synthetic and real-world datasets. We assess the performance of our fair graph construction methods by comparing it against traditional graph construction techniques. Additionally, we evaluate the effectiveness of the fair spectral clustering algorithm when applied to the constructed fair graphs. Furthermore, we compare the performance of the fair spectral clustering algorithm with that of the traditional spectral clustering algorithm using the fair graph construction method.

\subsection{Datasets}
We describe the datasets used in our experiments. To evaluate the performance of our methods, we utilized both synthetic and real-world datasets. The synthetic dataset was generated using the Stochastic Block Model (SBM) as proposed in \cite{Kleindessner_Samadi_Awasthi_Morgenstern_2019}. The real-world tabular datasets were sourced from the UCI Machine Learning repository and the datasets as described in \cite{Fabris2022AlgorithmicFD}. In addition, we employed several real-world image datasets to further assess the performance of our methods.

The details of the datasets are given below, starting with the synthetic dataset:

\textbf{Stochastic Block Model (SBM)}: We utilize a modified version of the SBM proposed in \cite{Kleindessner_Samadi_Awasthi_Morgenstern_2019}. In this model, a total of $n$ vertices are divided into $h$ disjoint sensitive groups, such that the vertex set $V = V_{1} \cup V_{2} \cup \dots \cup V_{h}$. Additionally, the vertices are assigned to $c$ clusters corresponding to a given ground truth fair clustering, i.e., $V = C_{1} \cup C_{2} \cup \dots \cup C_{c}$. The parameters for generating the data points are specified in Equation~\ref{eqn:SBM_parameters}. For all $g \in \{1 \dots h\}$ and $l \in \{1 \dots c\}$:

Number of vertices in a sensitive group:
\begin{equation}\label{eqn:SBM_parameters}
\left |V_{g} \right | = \frac{n}{h}
\end{equation}

Number of vertices in a cluster:
\begin{equation}
\left | C_{l} \right | = \frac{n}{c}
\end{equation}

Fraction of vertices from sensitive group $g$ in cluster $C_l$:
\begin{equation}
\frac{\left | V_{g} \cap C_{l} \right |}{\left | C_l \right |} = \frac{1}{h}
\end{equation}

Furthermore, edges between pairs of vertices are added probabilistically as follows:

\begin{equation*}
a_{ij} = \begin{cases}
p & \text{if } v_i,v_j \text{ in same cluster, same group} \\
q & \text{if } v_i,v_j \text{ in different clusters, same group} \\
r & \text{if } v_i,v_j \text{ in same cluster, different groups} \\
s & \text{if } v_i,v_j \text{ in different clusters, different groups}
\end{cases}
\end{equation*}

Here, $a_{ij}$ denotes the probability of an edge between nodes $i$ and $j$, with the edge probabilities satisfying $p > q > r > s$. Each edge is assigned a random weight from the interval $[0.1, 2]$, and the weights are normalized such that the sum of edge weights for each node equals 1.

Since our method relies on graph construction, we also assign features to each node. As described in \cite{zhang2024dual}, these features are sampled from a normal distribution:

\begin{equation}
\textbf{X}_{[:,d]} \sim \mathcal{N}(\textbf{0}, (\textbf{L})^{\dagger} + \boldsymbol{\xi})
\end{equation}

where $\boldsymbol{\xi}$ is a noise vector, $(\textbf{L})^{\dagger}$ is the pseudo-inverse of the graph Laplacian $\textbf{L}$, and $d$ denotes the number of features. Accordingly, we generate $n$ feature vectors, each of dimension $d$, drawn from a multivariate normal distribution with mean zero and covariance matrix $(\textbf{L})^{\dagger} + \boldsymbol{\xi}$. As noted in \cite{dong2016learning}, this feature generation process yields signals that are smooth across the graph.
For the experiments we follow a similar setup as in \cite{zhang2024dual}, with parameters such as number of features $d = 100$, number of clusters $c = 4$, number of sensitive groups $h = 2$, and varying number of data points $n = 1000$ and $n = 5000$.

We ran the experiments on seven real-world tabular datasets. Further, the results of those experiments were used to analyze the performance of our fair graph construction methods. The descriptions of real-world datasets we used are as given below:

\begin{table*}
\centering
\caption{Results on SBM dataset}
\label{table:synthetic_results}
\begin{threeparttable}
\begin{tabular}{|l|l|l|l|l|l|l|l|l|} 
\hline
                                  & \multicolumn{2}{c|}{uni. $\boldsymbol{\xi} \in$ [0, 0.2], n = 1000} & \multicolumn{2}{c|}{uni. $\boldsymbol{\xi} \in$ [0.4, 0.6], n = 1000} & \multicolumn{2}{c|}{uni. $\boldsymbol{\xi} \in$ [0, 0.2], n = 5000} & \multicolumn{2}{c|}{uni. $\boldsymbol{\xi} \in$ [0.4, 0.6], n = 5000}  \\ 
\hline
                                  & \textbf{CE}            & \textbf{Balance}                                  & \textbf{CE}            & \textbf{Balance}                                     & \textbf{CE}            & \textbf{Balance}                                  & \textbf{CE}            & \textbf{Balance}                                      \\ 
\hline
\textbf{kNN-SC}                   & 0.84          & 0.04                                     & 0.8           & 0.06                                        & 0.81          & 0.05                                     & 0.83          & 0.09                                         \\ 
\hline
\textbf{$\epsilon$-SC}               & 0.67          & 0.5                                      & 0.7           & 0.43                                        & 0.66          & 0.51                                     & 0.59          & 0.47                                         \\ 
\hline
\textbf{RGC*}                     & 0.72          & 0.18                                     & 0.71          & 0.15                                        & 0.83*         & 0.12*                                    & 0.55*         & 0.39*                                        \\ 
\hline
\textbf{SGL}                      & 0.64          & 0.36                                     & \textbf{0.51} & 0.59                                        & 0.67          & 0.30                                     & 0.69          & 0.31                                         \\ 
\hline
\textbf{CDC}                      & 0.88          & 0.20                                     & 0.89          & 0.11                                        & 0.75          & 0.19                                     & 0.77          & 0.26                                         \\ 
\hline
\textbf{SFC}                      & 0.88          & 0.20                                     & 0.89          & 0.11                                        & 0.75          & 0.19                                     & 0.77          & 0.26                                         \\ 
\hline
\textbf{kNN-FSC}                  & 0.58          & 0.57                                     & 0.68          & 0.43                                        & \uline{0.53}  & 0.63                                     & 0.67          & 0.66                                         \\ 
\hline
\textbf{$\epsilon$-FSC}              & 0.63          & 0.77                                     & 0.69          & 0.66                                        & 0.71          & 0.64                                     & 0.77          & 0.7                                          \\ 
\hline
\textbf{fair\_kNN-SC (Ours)}      & \textbf{0.52} & 0.88                                     & \uline{0.58}          & \textbf{0.85}                               & \textbf{0.52} & \textbf{0.88}                            & 0.54          & 0.79                                         \\ 
\hline
\textbf{fair\_$\epsilon$-SC (Ours)}  & \uline{0.56}  & \textbf{0.84}                            & \textbf{0.51} & \textbf{0.85}                               & \textbf{0.52} & \textbf{0.88}                            & \textbf{0.52} & \textbf{0.86}                                \\ 
\hline
\textbf{fair\_kNN-FSC (Ours)}     & \uline{0.56}  & \uline{0.81}                             & 0.62          & 0.79                                        & 0.55          & 0.81                            & 0.54          & 0.78                                         \\ 
\hline
\textbf{fair\_$\epsilon$-FSC (Ours)} & 0.59          & \textbf{0.84}                            & 0.59          & \uline{0.82}                                & \uline{0.53}  & \uline{0.83}                                     & \uline{0.53}  & \uline{0.85}                                 \\
\hline
\end{tabular}
\begin{tablenotes}
  \small
  \item  The best and second-best values are highlighted in bold and underline, respectively. *~denotes that for RGC the experiments could not be run for 5000 data points due to computational constraints. The results were obtained for 2000 data points.
\end{tablenotes}
\end{threeparttable}
\end{table*}

\begin{enumerate}
  \item \textbf{Adult Income Dataset} \cite{misc_adult_2} (Adult): This dataset includes demographic and employment-related information such as education, occupation, and age. In our analysis, we treated \textit{gender} as the sensitive attribute.
  
  \item \textbf{Banking Dataset} \cite{misc_bank_marketing_222} (Bank): This dataset contains data from marketing campaigns conducted by a Portuguese bank. It includes features such as age, loan status, and account balance. We designated \textit{marital status} as the sensitive attribute, focusing only on the categories ``married'' and ``single'', excluding ``divorced'' from our analysis.
  
  \item \textbf{Catalonia Juvenile Dataset} \cite{miron2021evaluating1} (Catalonia): This dataset provides information on juveniles who were part of the justice system in Catalonia. We considered the \textit{gender} of the individuals as the sensitive feature.
  
  \item \textbf{COMPAS Recidivism Dataset} \cite{githubGitHubPropublicacompasanalysis} (Compas): This dataset offers details on recidivism outcomes for individuals in the U.S. criminal justice system. The sensitive attribute used in our study was \textit{race}.
  
  \item \textbf{Default of Credit Card Clients Dataset} \cite{misc_default_of_credit_card_clients_350} (Credit): This dataset includes demographic and financial information related to credit card users in Taiwan from April to September 2005. We used \textit{gender} as the sensitive attribute for our experiments.
  
  \item \textbf{Crime and Communities Dataset} \cite{communities_and_crime_183} (Crime): This dataset contains socio-economic (e.g., income), demographic (e.g., race), and law enforcement-related (e.g., patrolling) information for neighborhoods across the United States. We considered the percentage of white population in a neighborhood as the sensitive attribute. The neighborhood with the majority white population was considered as the one sensitive group, and the neighborhood with the majority non-white population was considered as the other sensitive group.
  
  \item \textbf{Law Admission Dataset} \cite{wightman1998lsac} (Law): Based on a 1991 survey covering 163 U.S. law schools, this dataset includes admissions-related data. We identified \textit{race} as the sensitive attribute.
  
  \item \textbf{Student Performance Dataset} \cite{misc_student_performance_320} (Student): This dataset records academic performance in Mathematics and Portuguese along with demographic details. We used \textit{gender} as the sensitive attribute in our analysis.

\end{enumerate}

We also used three real-world image datasets to evaluate the performance of our methods. The descriptions of these datasets are as given below:

\begin{enumerate}
  \item \textbf{MNIST-USPS} \cite{li2020deep}: This dataset is a combination of the MNIST and USPS datasets. It contains images of handwritten digits from 0 to 9. The sensitive attribute is the dataset from which the image is taken, i.e., MNIST or USPS.
  \item \textbf{Reverse MNIST} \cite{li2020deep}: This dataset is a combination of the MNIST and Reverse MNIST datasets. It contains images of handwritten digits from 0 to 9. The sensitive attribute is the dataset from which the image is taken, i.e., MNIST or Reverse MNIST.
  \item \textbf{MTFL} \cite{li2020deep}: This dataset consists of 12,995 images used for facial recognition. The labels contain gender, with or without glasses, smiling, etc. We considered wearing glasses as the sensitive attribute. 
\end{enumerate}

\begin{table*}
\centering
\begin{threeparttable}
  \caption{Results on real-world tabular datasets}
\label{table:real_world_results}
\begin{tabular}{|c|ccc|ccc|ccc|ccc|} 
\hline
\textbf{Method}                                                     & \multicolumn{3}{c|}{\textbf{Adult}}                & \multicolumn{3}{c|}{\textbf{Bank}}                 & \multicolumn{3}{c|}{\textbf{Catalonia}}            & \multicolumn{3}{c|}{\textbf{Compas}}                \\ 
\hline
                                                                    & \textbf{Balance} & \textbf{NCut}  & \textbf{SS}    & \textbf{Balance} & \textbf{NCut}  & \textbf{SS}    & \textbf{Balance} & \textbf{NCut}  & \textbf{SS}    & \textbf{Balance} & \textbf{NCut}  & \textbf{SS}     \\ 
\hline
\textbf{kNN-SC}                                                     & 0.157            & 0.516          & 0.440           & 0.219            & 0.512          & 0.413          & 0.159            & 0.223          & 0.521          & 0.362            & 0.278          & 0.467           \\ 
\hline
\textbf{$\epsilon$-SC}                                                 & 0.092            & 0.511          & 0.440           & 0.021            & 0.555          & 0.448          & 0.083            & 0.387          & 0.498          & 0.193            & 0.312          & 0.523           \\ 
\hline
\textbf{RGC}                                                        & 0.193            & 0.522          & 0.420           & 0.330             & 0.587          & 0.447          & 0.203            & 0.334          & 0.512          & 0.517            & 0.291          & 0.478           \\ 
\hline
\textbf{SGL}                                                        & 0.274            & 0.571          & 0.431          & 0.230             & 0.551          & 0.448          & 0.198            & 0.398          & 0.534          & 0.280             & 0.323          & 0.501           \\ 
\hline
\textbf{CDC}                                                        & 0.318            & 0.561          & 0.410           & 0.182            & 0.569          & 0.447          & 0.119            & 0.312          & 0.489          & 0.229            & 0.287          & 0.445           \\ 
\hline
\rowcolor[rgb]{0.871,0.867,0.855} \textbf{kNN-FSC}                  & 0.429            & \uline{0.819}  & \uline{0.412}  & 0.438            & \uline{0.644}  & 0.395          & \textbf{0.214}   & 0.445          & 0.389          & 0.600              & \textbf{0.467}          & 0.356           \\ 
\hline
\rowcolor[rgb]{0.871,0.867,0.855} \textbf{$\epsilon$-FSC}              & 0.333            & 0.871          & 0.320           & 0.301            & 0.724          & 0.294          & 0.192            & 0.478          & 0.342          & 0.501            & \uline{0.489}          & 0.331           \\ 
\hline
\rowcolor[rgb]{0.871,0.867,0.855} \textbf{SFC}                      & 0.412            & 0.933          & 0.320           & 0.442            & 0.83           & 0.331          & 0.188            & 0.521          & 0.298          & 0.612            & 0.534          & 0.287           \\ 
\hline
\rowcolor[rgb]{0.871,0.867,0.855} \textbf{fair\_kNN-SC (Ours)}      & 0.471            & 0.914          & 0.340           & \uline{0.449}    & 0.799          & \uline{0.410}  & 0.209            & 0.498          & 0.367          & 0.656            & 0.521          & 0.334           \\ 
\hline
\rowcolor[rgb]{0.871,0.867,0.855} \textbf{fair\_$\epsilon$-SC (Ours)}  & 0.471            & \textbf{0.743} & \textbf{0.421} & 0.445            & \textbf{0.617} & \textbf{0.433} & \uline{0.211}    & \textbf{0.467} & \textbf{0.398} & \uline{0.668}    & \uline{0.489} & \textbf{0.367}  \\ 
\hline
\rowcolor[rgb]{0.871,0.867,0.855} \textbf{fair\_kNN-FSC (Ours)}     & \textbf{0.476}   & 0.932          & 0.330           & \textbf{0.467}   & 0.803          & 0.313          & \textbf{0.214}   & 0.523          & 0.334          & \textbf{0.669}   & 0.545          & 0.312           \\ 
\hline
\rowcolor[rgb]{0.871,0.867,0.855} \textbf{fair\_$\epsilon$-FSC (Ours)} & \uline{0.473}    & 0.952          & 0.390           & 0.445            & 0.702          & 0.292          & 0.204            & \uline{0.478}  & \uline{0.378}  & 0.667            & 0.498  & \uline{0.345}   \\ 
\hline
\rowcolor[rgb]{0.753,0.753,0.753} \textbf{Target\_Balance}          & 0.479            & —              & —              & 0.469            & —              & —              & 0.216            & —              & —              & 0.670             & —              & —               \\ 
\hline
\multicolumn{13}{c} {}                  \\ [-0.8em]
\hline
\textbf{Method}                                                                     & \multicolumn{3}{c|}{\textbf{Credit}}               & \multicolumn{3}{c|}{\textbf{Crime}}                & \multicolumn{3}{c|}{\textbf{Law}}                  & \multicolumn{3}{c|}{\textbf{Student}}               \\ 
\hline
                                                                    & \textbf{Balance} & \textbf{NCut}  & \textbf{SS}    & \textbf{Balance} & \textbf{NCut}  & \textbf{SS}    & \textbf{Balance} & \textbf{NCut}  & \textbf{SS}    & \textbf{Balance} & \textbf{NCut}  & \textbf{SS}     \\ 
\hline
\textbf{kNN-SC}                                                     & 0.553            & 0.334          & 0.460          & 0.023            & 0.281          & 0.546          & 0.098            & 0.251          & 0.456          & 0.599            & 0.201          & 0.511           \\ 
\hline
\textbf{$\epsilon$-SC}                                                 & 0.418            & 0.275          & 0.500          & 0.100            & 0.292          & 0.511          & 0.045            & 0.265          & 0.474          & 0.343            & 0.287          & 0.498           \\ 
\hline
\textbf{RGC}                                                        & 0.166            & 0.259          & 0.511          & 0.061            & 0.251          & 0.456          & 0.012            & 0.253          & 0.489          & 0.300              & 0.245          & 0.452           \\ 
\hline
\textbf{SGL}                                                        & 0.318            & 0.210          & 0.529          & 0.068            & 0.345          & 0.545          & 0.101            & 0.337          & 0.526          & 0.591            & 0.259          & 0.506           \\ 
\hline
\textbf{CDC}                                                        & 0.227            & 0.209          & 0.546          & 0.082            & 0.268          & 0.549          & 0.133            & 0.349          & 0.515          & 0.140             & 0.304          & 0.511           \\ 
\hline
\rowcolor[rgb]{0.871,0.867,0.855} \textbf{kNN-FSC}                  & 0.620             & 0.539          & 0.352          & 0.162            & 0.527          & 0.469          & \textbf{0.181}   & 0.483          & 0.345          & 0.814            & 0.394          & 0.417           \\ 
\hline
\rowcolor[rgb]{0.871,0.867,0.855} \textbf{$\epsilon$-FSC}              & 0.509            & 0.472          & 0.362          & 0.101            & \uline{0.452}  & 0.471          & 0.132            & 0.477          & 0.300            & 0.727            & 0.396          & 0.458           \\ 
\hline
\rowcolor[rgb]{0.871,0.867,0.855} \textbf{SFC}                      & 0.563            & \uline{0.443}  & \uline{0.451}          & 0.143            & 0.453          & 0.328          & \uline{0.179}    & \textbf{0.322} & \uline{0.421}  & \uline{0.861}    & 0.390          & 0.475           \\ 
\hline
\rowcolor[rgb]{0.871,0.867,0.855} \textbf{fair\_kNN-SC (Ours)}      & \textbf{0.656}   & 0.504          & 0.327          & \textbf{0.189}            & 0.583          & 0.334          & \uline{0.179}    & 0.422          & 0.302          & \textbf{0.871}   & 0.418          & 0.354           \\ 
\hline
\rowcolor[rgb]{0.871,0.867,0.855} \textbf{fair\_$\epsilon$-SC (Ours)}  & 0.651            & \textbf{0.442} & \textbf{0.463} & 0.176    & \textbf{0.451} & \textbf{0.483} & 0.173            & \uline{0.377}  & \textbf{0.428} & 0.805            & \textbf{0.294} & \textbf{0.485}  \\ 
\hline
\rowcolor[rgb]{0.871,0.867,0.855} \textbf{fair\_kNN-FSC (Ours)}     & 0.642            & 0.495          & 0.426          & \uline{0.188}   & 0.647          & 0.347          & \uline{0.179}    & 0.419          & 0.366          & \uline{0.861}    & 0.450          & 0.380           \\ 
\hline
\rowcolor[rgb]{0.871,0.867,0.855} \textbf{fair\_$\epsilon$-FSC (Ours)} & \uline{0.654}    & \uline{0.443}  & \textbf{0.463}  & 0.113            & 0.481          & \uline{0.476}  & 0.173            & 0.456          & 0.378          & 0.820            & \uline{0.296}  & \uline{0.484}   \\ 
\hline
\rowcolor[rgb]{0.753,0.753,0.753} \textbf{Target\_Balance}          & 0.656            & —              & —              & 0.192            & —              & —              & 0.189            & —              & —              & 0.899            & —              & —               \\
\hline
\end{tabular}
\begin{tablenotes}
  \small
  \item  The Balance, NCut, and SS metrics are calculated for the clusters obtained. The Balance for the total dataset is given in the last row. The fair methods are given a light grey background and the target Balance is given a dark grey background. The best and second-best values in fair methods are highlighted in bold and underline, respectively.
\end{tablenotes}
\end{threeparttable}
\end{table*}

\subsection{Metrics}

We considered the following metrics for evaluating the proposed fair graph construction methods:
\begin{itemize}
\item \textbf{Clustering Error (CE)}: This metric quantifies clustering performance by computing the proportion of incorrectly assigned labels relative to the total number of data points.
\begin{equation}
\text{CE} = \frac{1}{n} \sum_{i=1}^{n} \textbf{1}_{\textbf{y}_{i} \neq \textbf{y}_{i}^{*}}
\end{equation}
Here, $\textbf{y}_{i}$ denotes the predicted label for the $i$th sample, while $\textbf{y}_{i}^{*}$ represents its ground truth label. This is calculated specifically for synthetic datasets, as the true labels correspond to the ground truth fair clustering. 

\item \textbf{Balance}:  
Balance is a widely used and algorithm agnostic group-level fairness metric in clustering. For binary sensitive groups, say group 1 and group 2, the balance of a cluster \(C\) is defined as:

\begin{equation}
  \mathrm{Balance}(C) \;=\; \min\!\left(\frac{|C^{1}|}{|C^{2}|},\;\frac{|C^{2}|}{|C^{1}|}\right)
\end{equation}
where \(|C^{1}|\) and \(|C^{2}|\) denote the counts of points from each sensitive group within cluster \(C\).  
The overall clustering balance is the minimum of these values across all clusters:
\begin{equation}
\mathrm{Balance}(\mathcal{C}) = \min_{C\in\mathcal{C}} \mathrm{Balance}(C)
\end{equation}
This metric ranges from 0 (completely unbalanced, e.g., a cluster with only one group) to 1 (perfectly balanced, equal group sizes in each cluster). A higher balance score indicates a fairer clustering outcome.

\item \textbf{Clustering Accuracy (Acc)}: The clustering accuracy is calculated as the ratio of the number of correctly classified points to the total number of points. This is calculated as follows:
\begin{equation}
\text{Acc} = \frac{1}{n} \sum_{i=1}^{n} \textbf{1}_{\textbf{y}_{i} = \textbf{y}_{i}^{*}}
\end{equation}
where $\textbf{y}_{i}$ is the predicted label of the $i$th point and $\textbf{y}_{i}^{*}$ is the true label of the $i$th point. We do not calculate accuracy for real-world tabular datasets as in most of the cases, the true labels in itself is not fair. 
\item \textbf{Silhouette Score (SS)}: The silhouette score is a measure of how similar an object is to its own cluster compared to other clusters. It is calculated as follows:

\begin{equation}
s_{i} = \frac{b_{i} - a_{i}}{\max(a_{i}, b_{i})}
\end{equation}

where $a_{i}$ is the average distance between the $i$th point and all other points in the same cluster, and $b_{i}$ is the average distance between the $i$th point and all points in the nearest cluster.

\begin{equation}
\text{SS} = \frac{1}{n} \sum_{i=1}^{n} s_{i}
\end{equation}

The silhouette score takes values between -1 and 1, where a higher value indicates better clustering.
\item \textbf{NCut Value (NCut)}: The normalized cut value is a measure of the quality of the clustering. It evaluates the total disconnection between clusters relative to the size of each cluster.

First, for each cluster $C_i$, compute the ratio:

\begin{equation}
\text{NCut}_i = \frac{\text{cut}(C_i, \bar{C_i})}{\text{vol}(C_i)} = \frac{\sum_{j \in C_i} \sum_{k \in \bar{C_i}} w_{jk}}{\sum_{j \in C_i} d_j}
\end{equation}

In this formulation, $\text{cut}(C_i, \bar{C_i})$ represents the total weight of the edges connecting the nodes in the cluster $C_i$ to the nodes outside of it, denoted as $\bar{C_i}$. The term $\text{vol}(C_i)$ refers to the sum of the degrees of all nodes within the cluster $C_i$. The weight of an edge between nodes $j$ and $k$ is represented by $w_{jk}$. Finally, the degree of a node $j$, denoted by $d_j$, is defined as the sum of the weights of all edges connected to it, i.e., $d_j = \sum_{k} w_{jk}$.

Then, the overall normalized cut is the sum over all the clusters:

\begin{equation}
\text{NCut} = \sum_{i=1}^{c} \text{NCut}_i
\end{equation}

The normalized cut value ranges between 0 and 1, where a lower value indicates better clustering, meaning stronger intra-cluster connections and weaker inter-cluster connections.
\end{itemize}

\subsection{Experimental Parameters}
In kNN graph construction, we set the number of neighbors $k = \sqrt{n}$, where $n$ is the number of data points. In $\epsilon$-neighborhood graph construction, we set $\epsilon = (log(n)/n)^{d}$.

To check if the choice of $k$ and $\epsilon$ can significantly affect the performance of the graph construction methods, we conducted experiments with different values of $k$ and $\epsilon$. The main parameter present in our method is the disparate impact parameter $\alpha$. In experiments where $\alpha$ is set as fixed value, we set $\alpha = 0.8$. We also experimented with different values of $\alpha$ to see how it affects the performance of the method. For the density reachable points, we set the minimum number of points to be 3. This was followed from the experiments of \cite{dbscan}, where they set the minimum number of points to be 3 for the density reachable points.

\subsection{Baseline Comparison Methods}
We compared the performance of our fair graph construction method with the following baseline methods:
\begin{itemize}
\item \textbf{kNN-SC}: Spectral clustering (SC) with kNN graph construction.
\item \textbf{$\epsilon$-SC}: Spectral clustering with $\epsilon$-neighborhood graph construction.
\item \textbf{kNN-FSC}: Fair spectral clustering (FSC) with kNN graph construction.
\item \textbf{$\epsilon$-FSC}: Fair spectral clustering with $\epsilon$-neighborhood graph construction.
\item \textbf{RGC}: The implementation\footnote{https://github.com/sckangz/RGC} of \cite{Kang2018RobustGL} which was clustered using Spectral Clustering.
\item \textbf{SGL}: The implementation\footnote{https://github.com/sckangz/SGL} of \cite{Kang2021StructuredGL}. This method uses a k-means clustering algorithm on the constructed graph, which was followed here for the comparison.
\item \textbf{CDC}: The implementation\footnote{https://github.com/XieXuanting/CDC} of \cite{Kang2024CDCAS}. This method returns the clustering results directly.
\item \textbf{SFC}: The implementation\footnote{https://github.com/imtiazziko/Variational-Fair-Clustering} of \cite{ziko2021variational}. This method returns the clustering results directly.
\end{itemize}

Our proposed methods are compared with these baseline methods in two ways: (1) Spectral clustering (SC) combined with our fair graph construction methods, and (2) Fair spectral clustering (FSC) combined with our fair graph construction methods. The first method is referred to as fair\_kNN-SC and fair\_$\epsilon$-SC, and the second method is referred to as fair\_kNN-FSC and fair\_$\epsilon$-FSC, in the current research.

\begin{figure}[]
  \centering
    \includegraphics[width=0.3\textwidth]{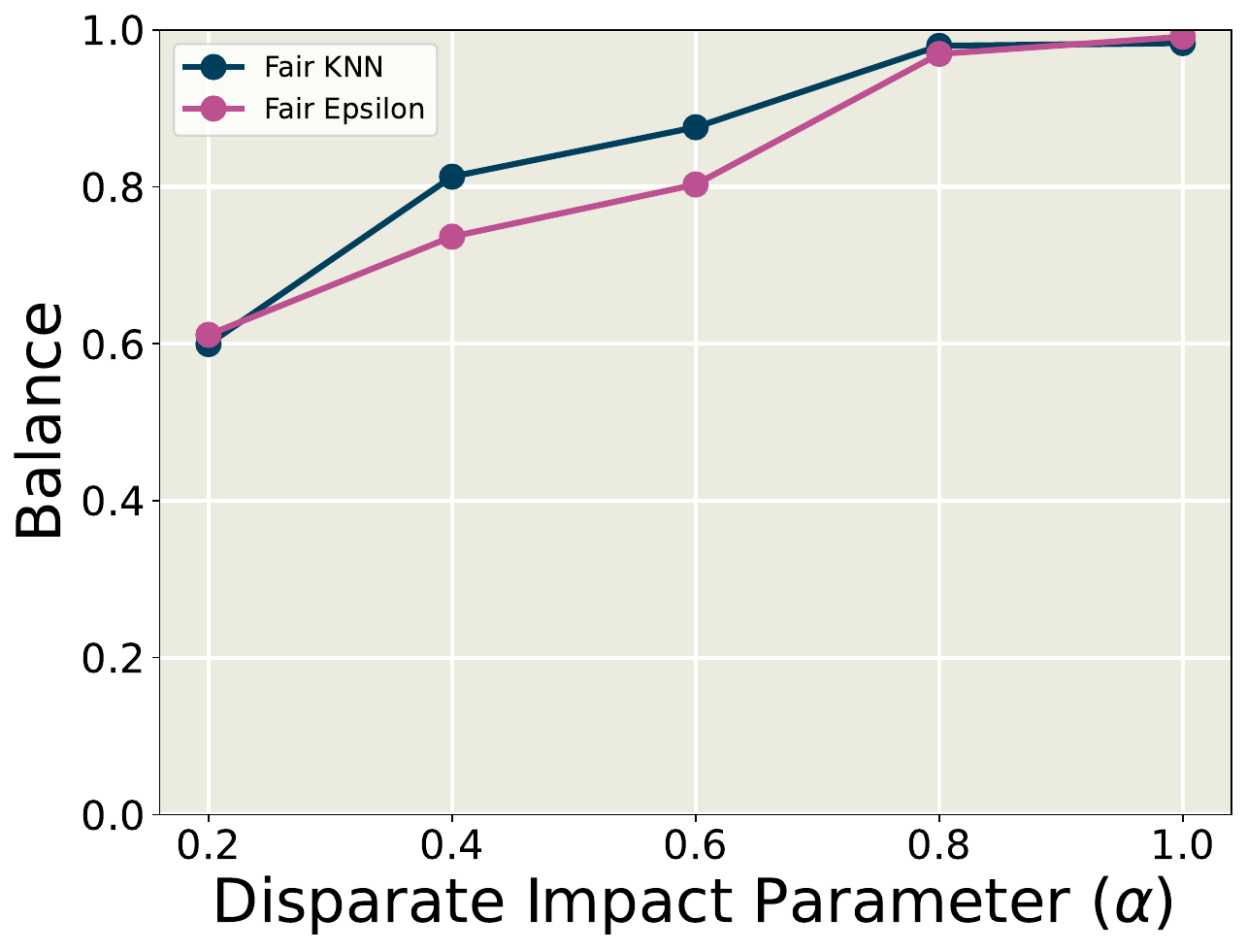}
    \caption{Change in Balance with the Disparate Impact parameter $\alpha$ in synthetic dataset}
    \label{fig:DI_Balance}
\end{figure}

\begin{figure*}[]
  \centering
  \begin{subfigure}[b]{0.24\textwidth}
    \includegraphics[width=\textwidth]{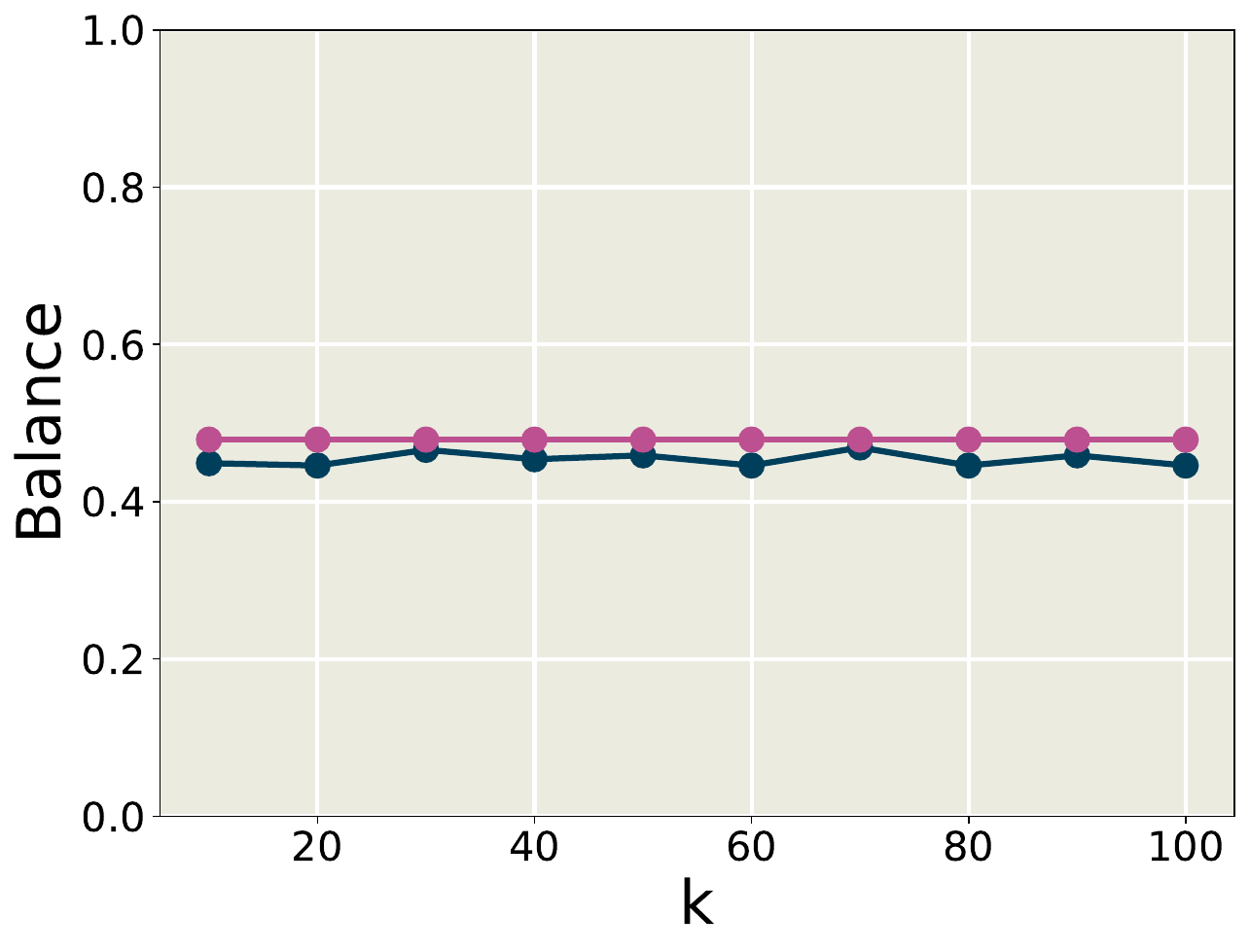}
    \caption{}
    \label{fig:Adult_balance_vs_k}
  \end{subfigure}
  \begin{subfigure}[b]{0.24\textwidth}
    \includegraphics[width=\textwidth]{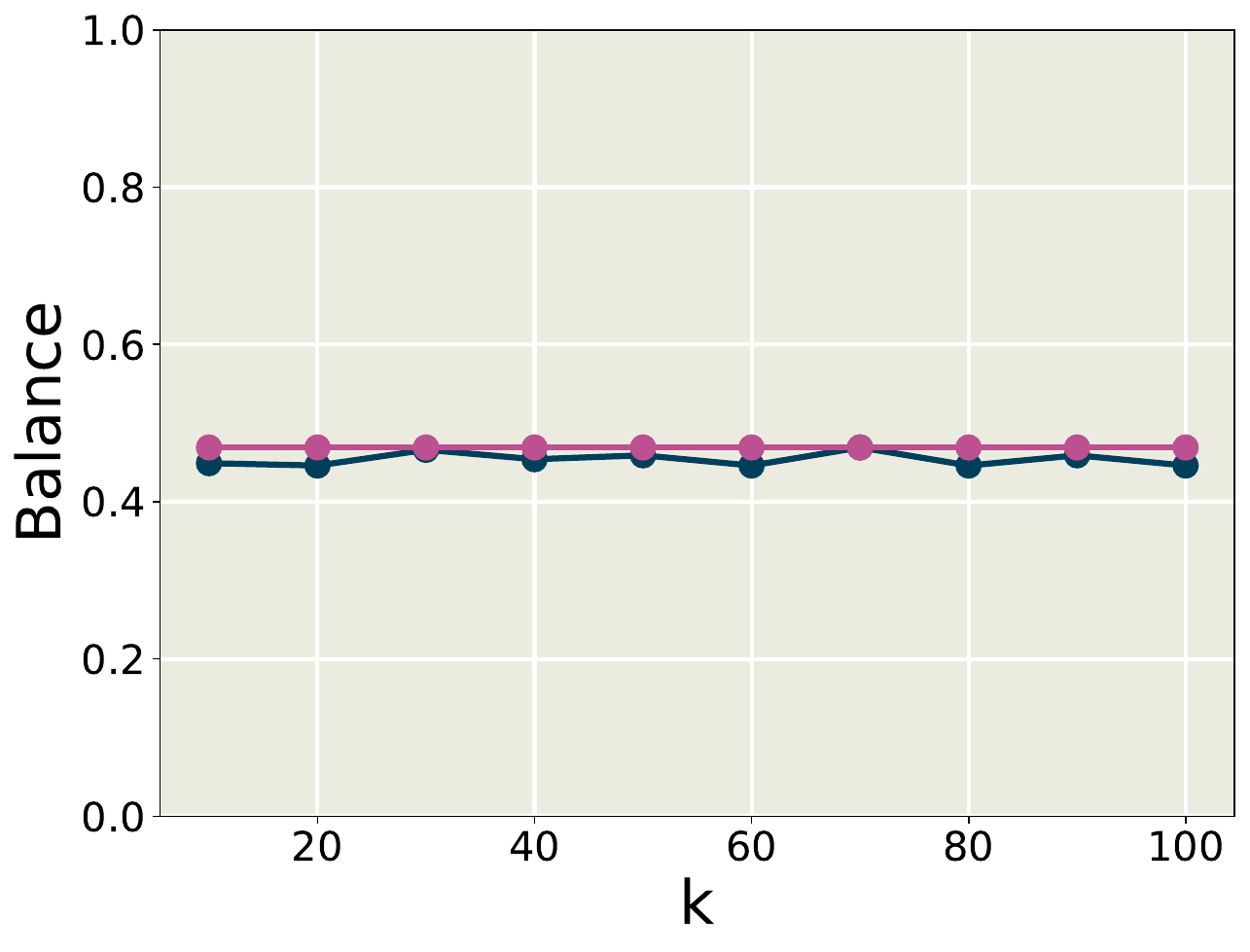}
    \caption{}
    \label{fig:Bank_balance_vs_k}
  \end{subfigure}
   \begin{subfigure}[b]{0.24\textwidth}
    \includegraphics[width=\textwidth]{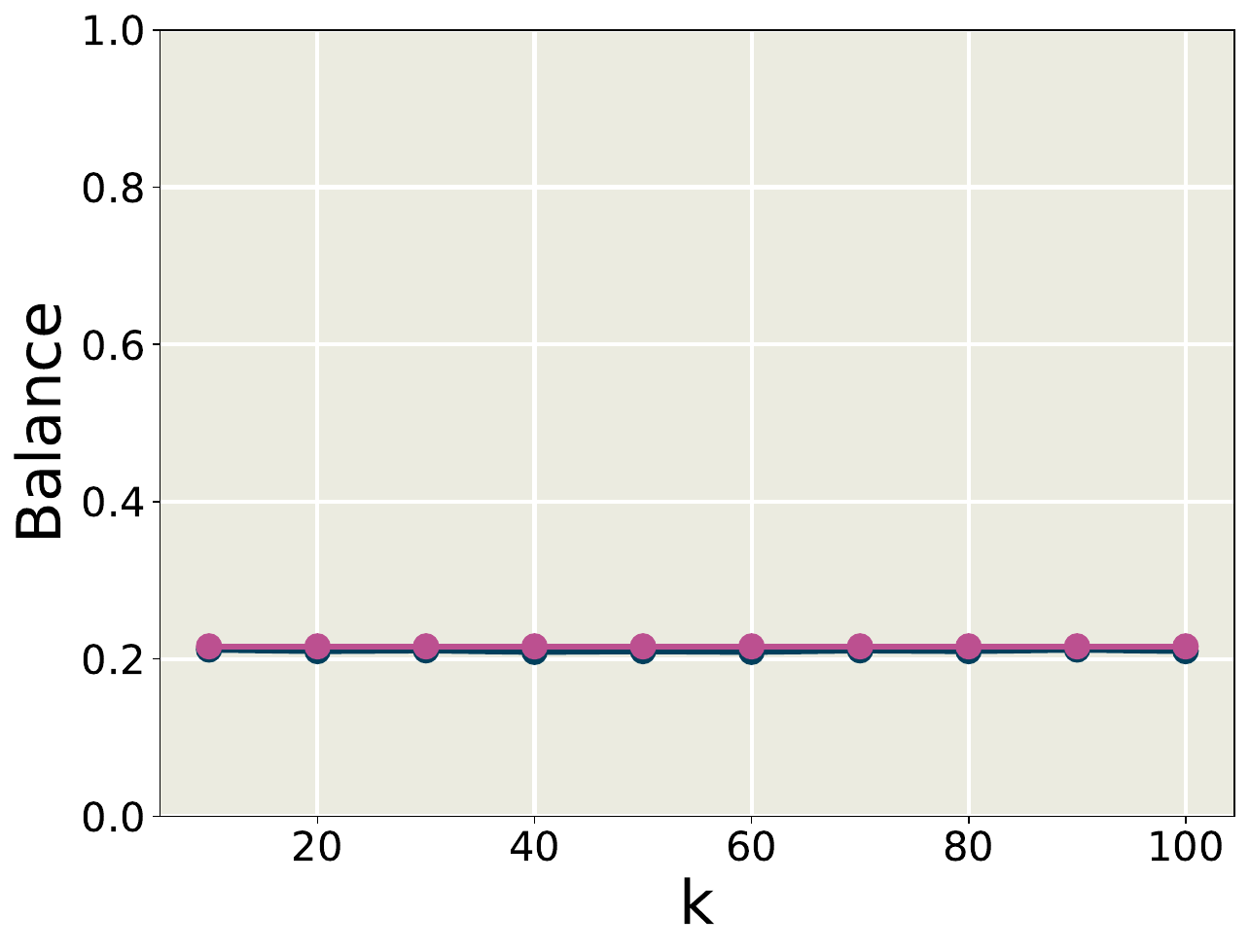}
    \caption{}
    \label{fig:Catalonia_balance_vs_k}
  \end{subfigure}
    \begin{subfigure}[b]{0.24\textwidth}
    \includegraphics[width=\textwidth]{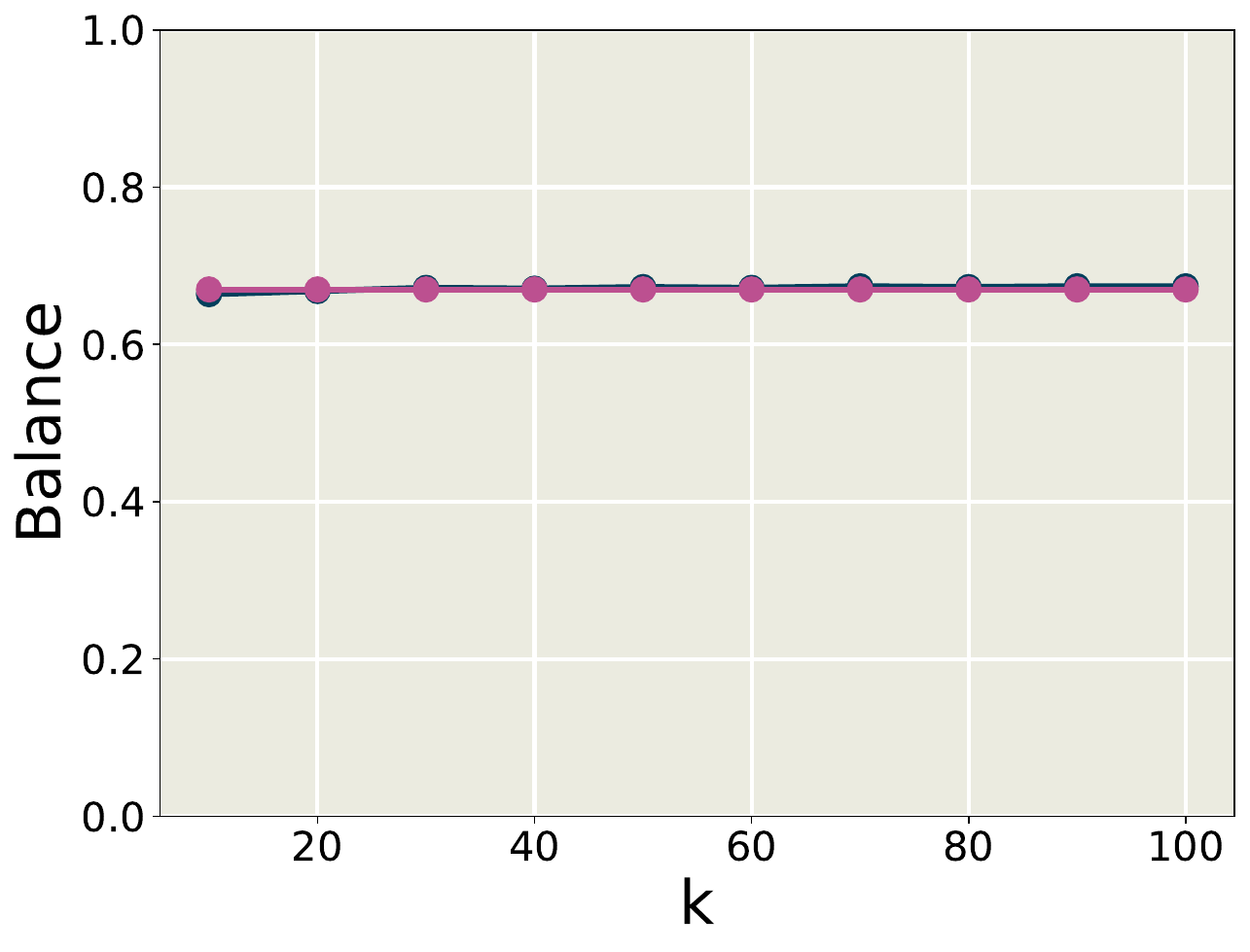}
    \caption{}
    \label{fig:Compas_balance_vs_k}
  \end{subfigure}
  \begin{subfigure}[b]{0.24\textwidth}
    \includegraphics[width=\textwidth]{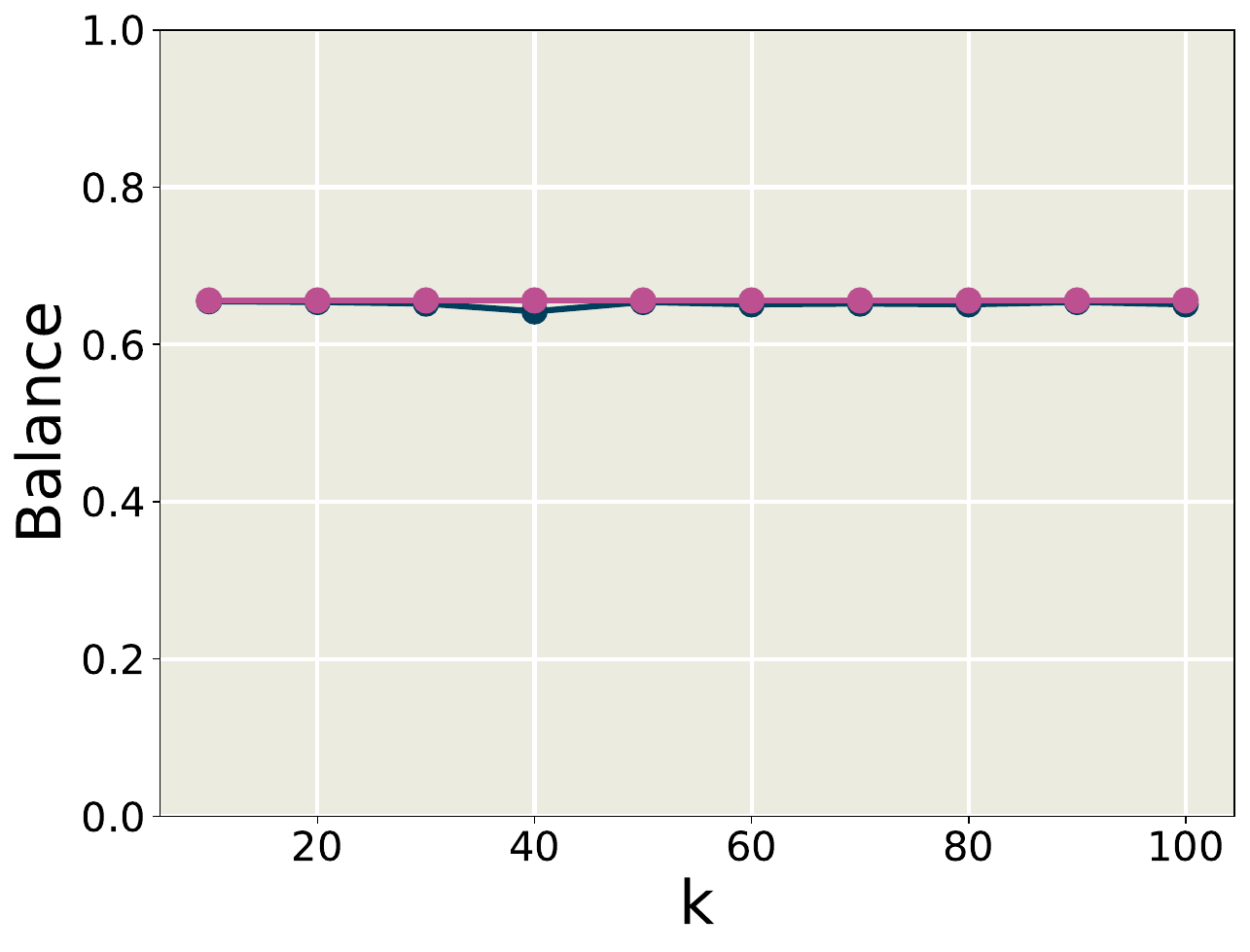}
    \caption{}
    \label{fig:Credit_balance_vs_k}
  \end{subfigure}
    \begin{subfigure}[b]{0.24\textwidth}
    \includegraphics[width=\textwidth]{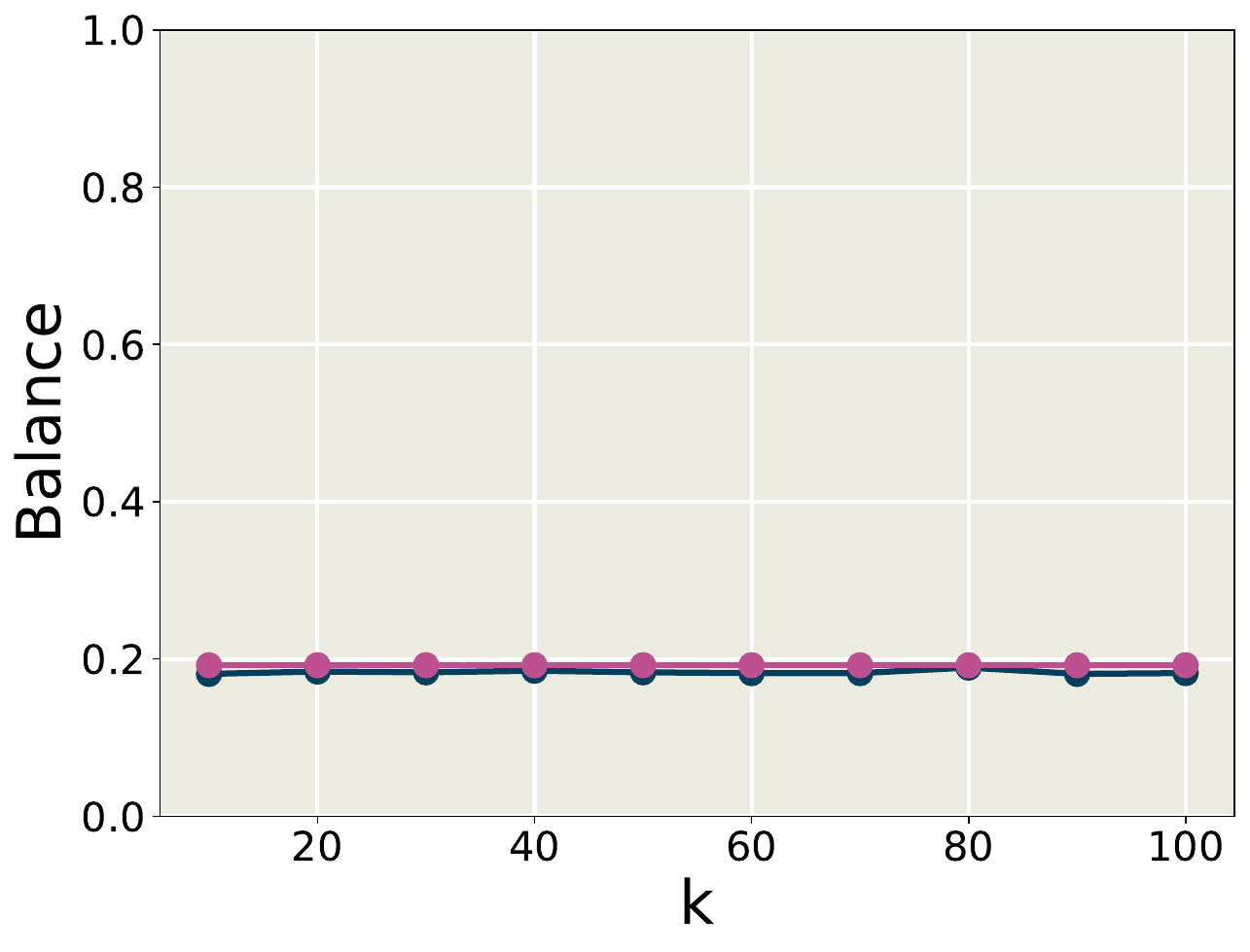}
    \caption{}
    \label{fig:Crime_balance_vs_k}
  \end{subfigure}
  \begin{subfigure}[b]{0.24\textwidth}
    \includegraphics[width=\textwidth]{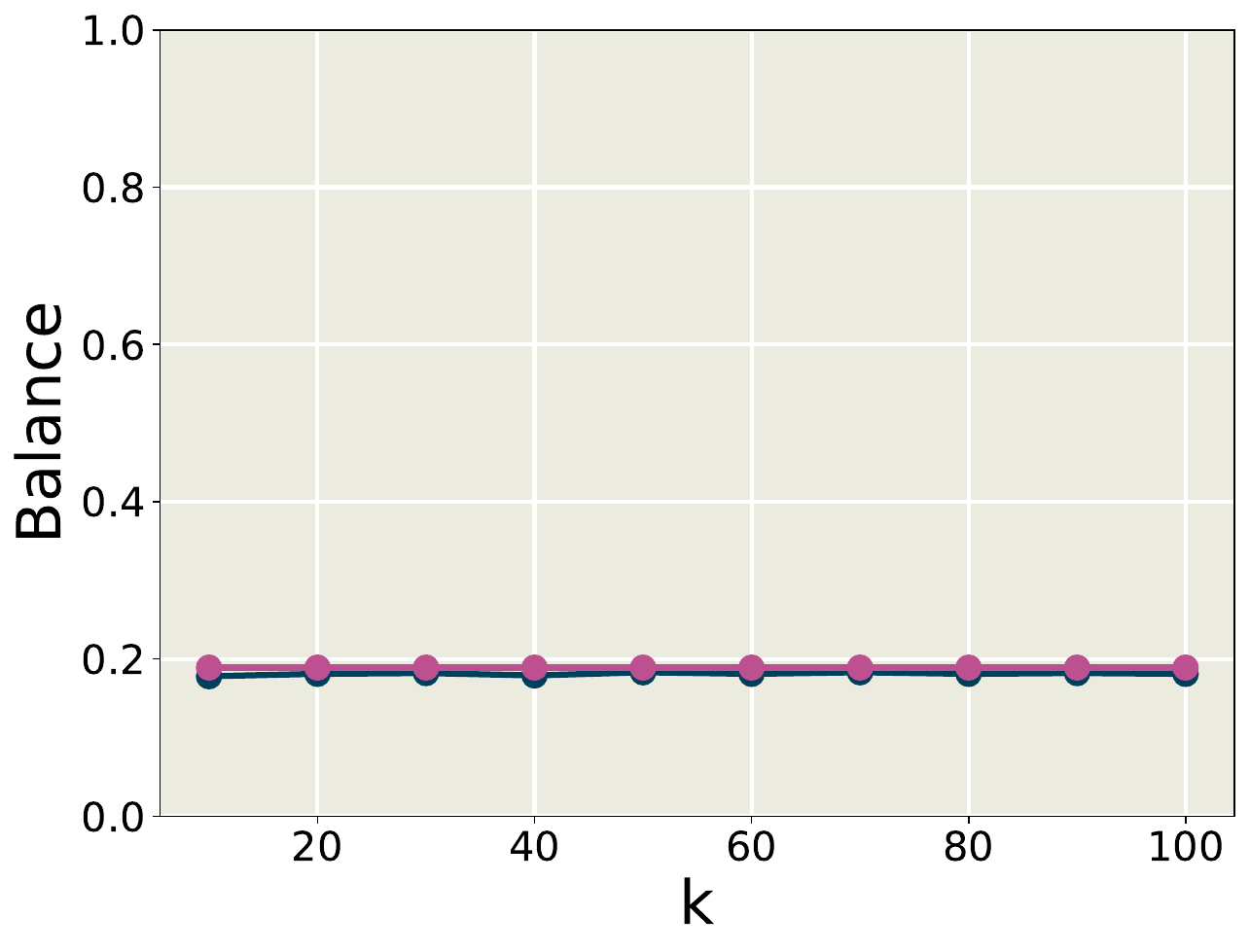}
    \caption{}
    \label{fig:Law_balance_vs_k}
  \end{subfigure}
  \begin{subfigure}[b]{0.24\textwidth}
    \includegraphics[width=\textwidth]{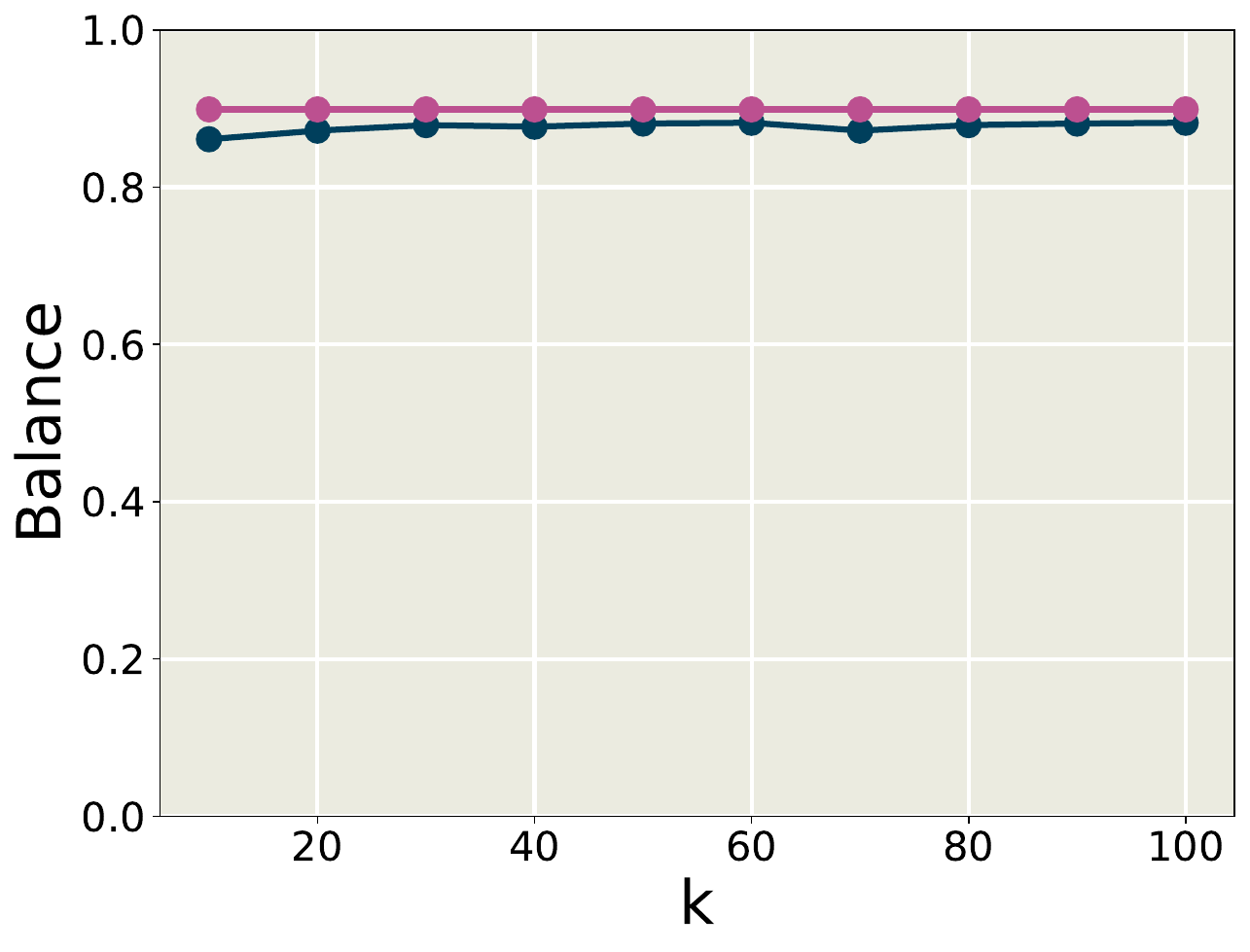}
    \caption{}
    \label{fig:Student_balance_vs_k}
  \end{subfigure}
  \begin{subfigure}[b]{0.24\textwidth}
    \includegraphics[width=\textwidth]{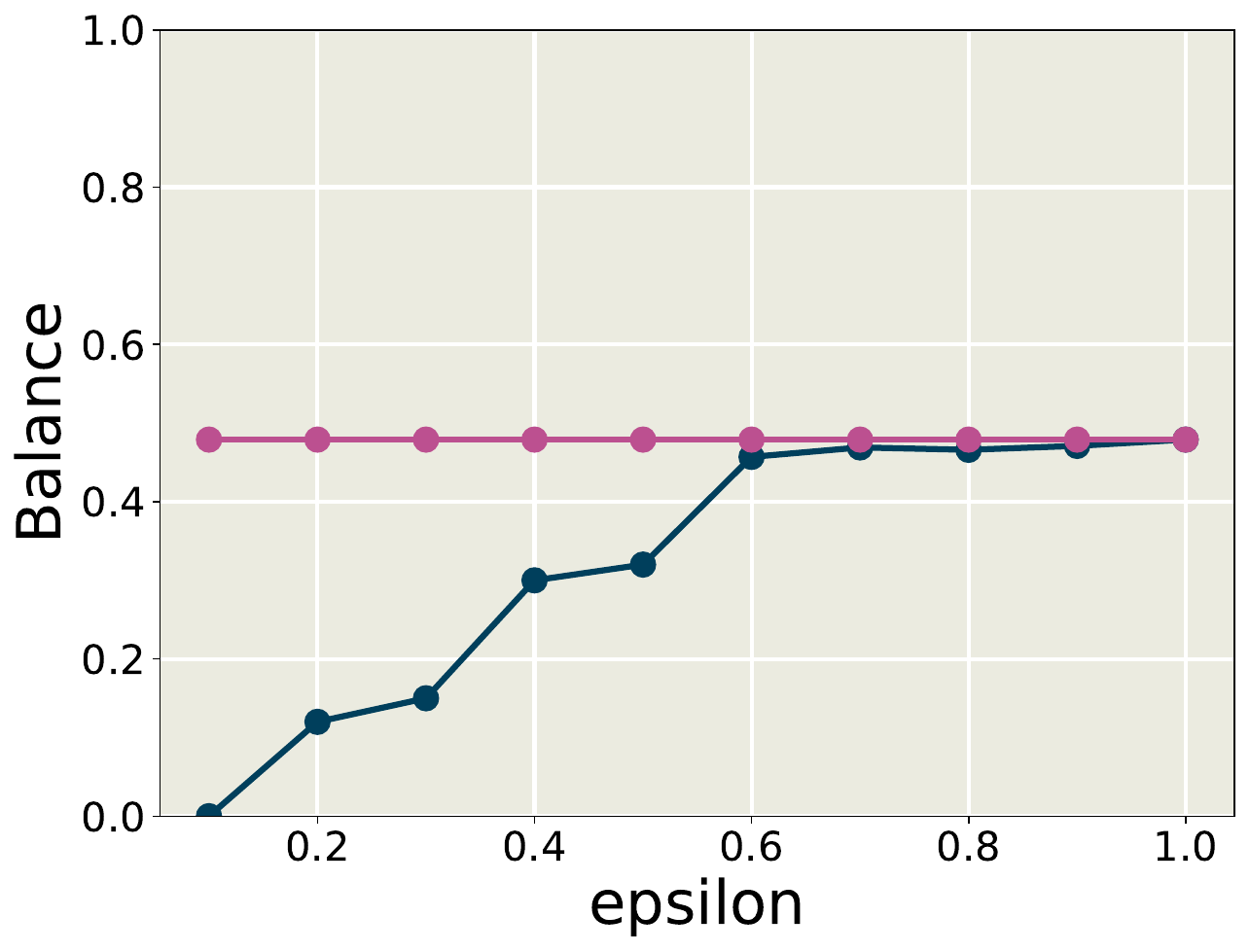}
    \caption{}
    \label{fig:Adult_balance_vs_eps}
  \end{subfigure}
  \begin{subfigure}[b]{0.24\textwidth}
    \includegraphics[width=\textwidth]{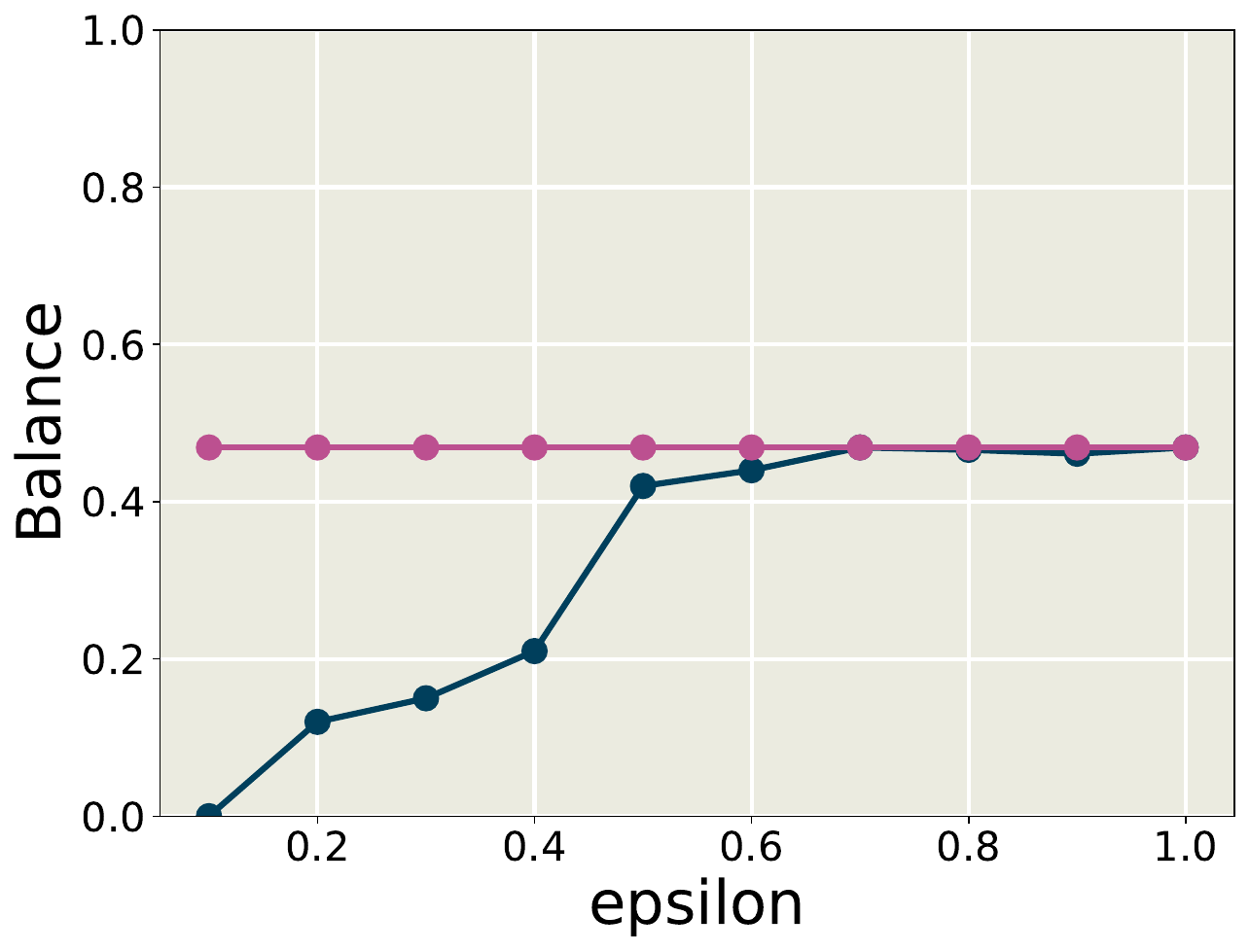}
    \caption{}
    \label{fig:Bank_balance_vs_eps}
  \end{subfigure}
    \begin{subfigure}[b]{0.24\textwidth}
    \includegraphics[width=\textwidth]{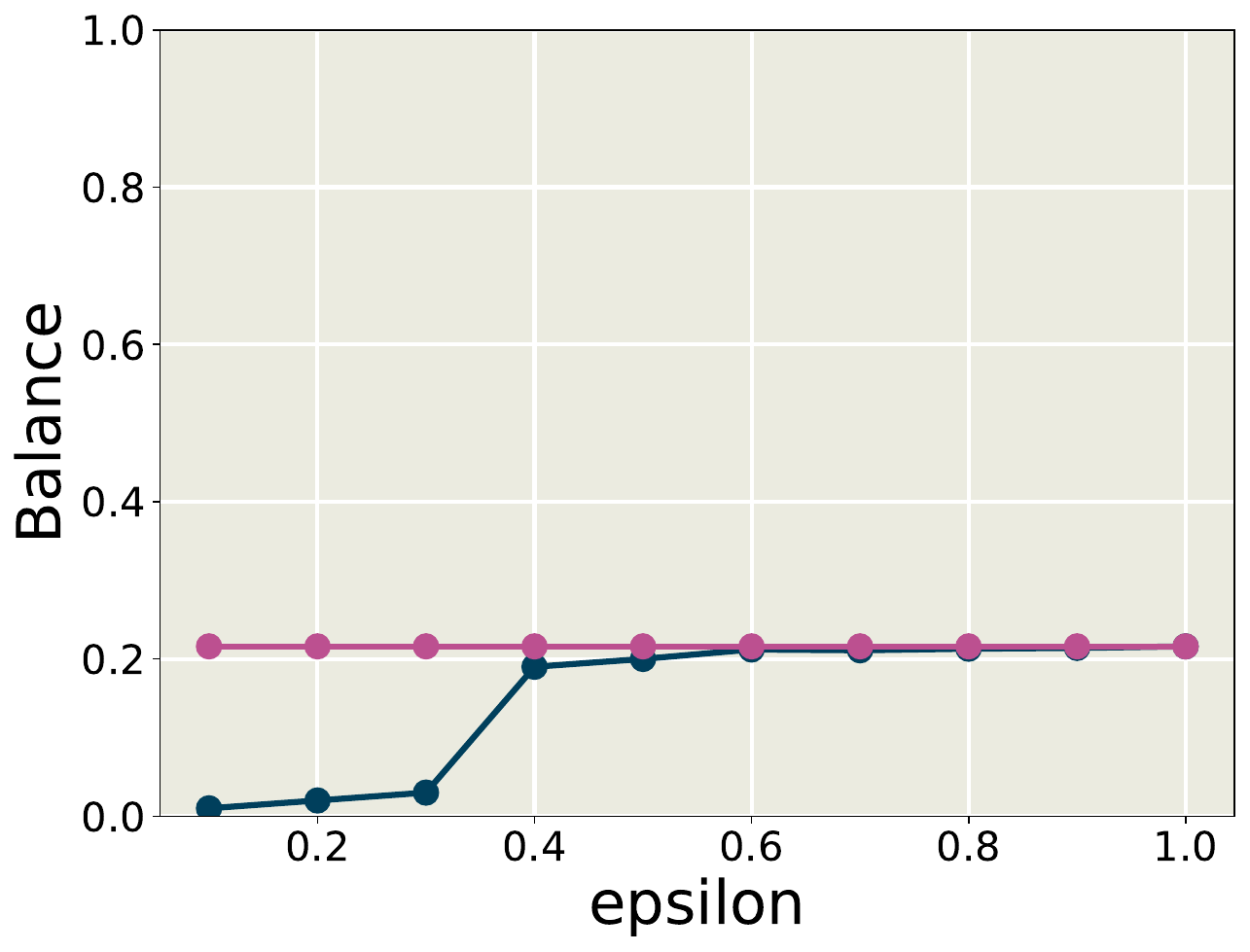}
    \caption{}
    \label{fig:Catalonia_balance_vs_eps}
  \end{subfigure}
    \begin{subfigure}[b]{0.24\textwidth}
    \includegraphics[width=\textwidth]{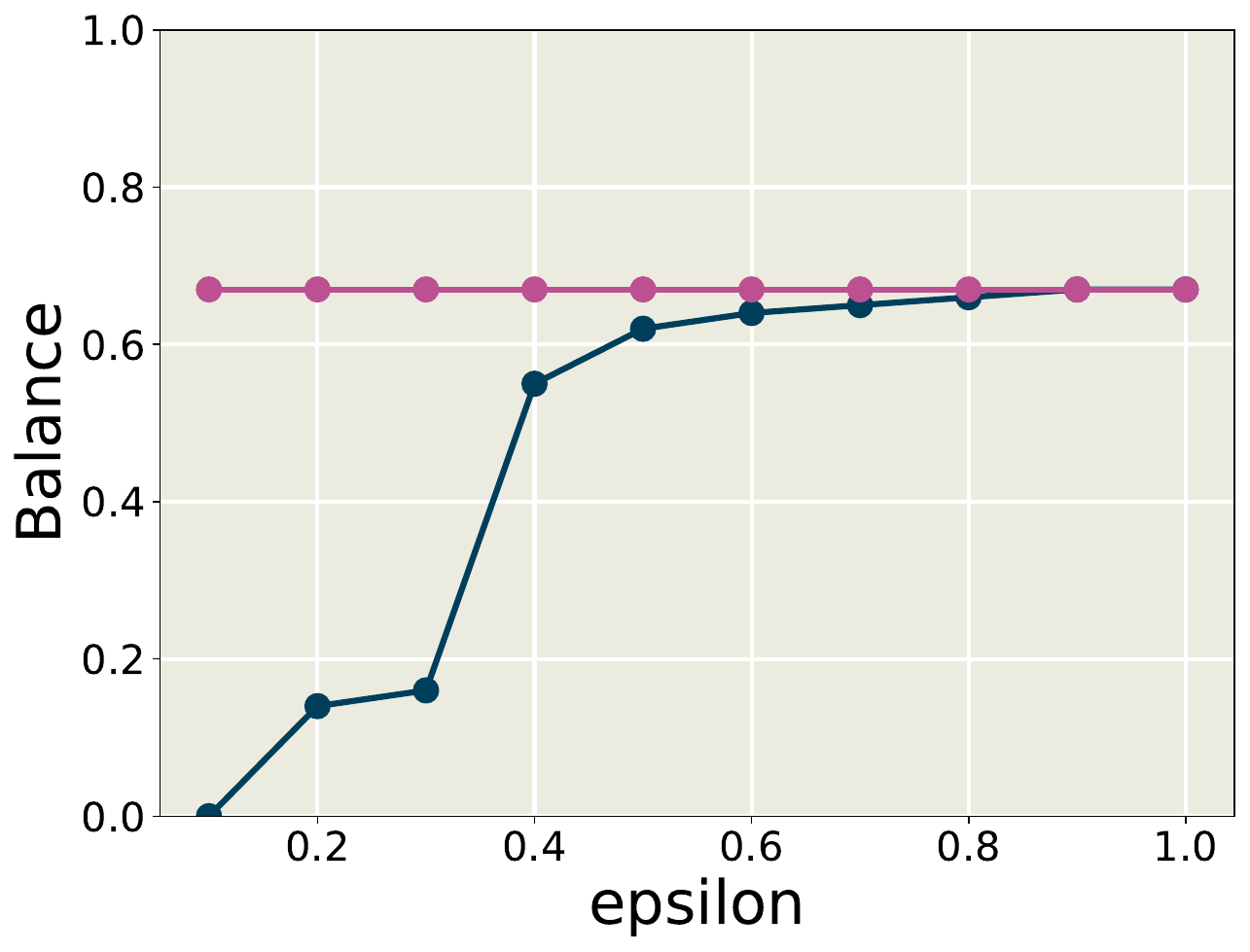}
    \caption{}
    \label{fig:Compas_balance_vs_eps}
  \end{subfigure}
  \begin{subfigure}[b]{0.24\textwidth}
    \includegraphics[width=\textwidth]{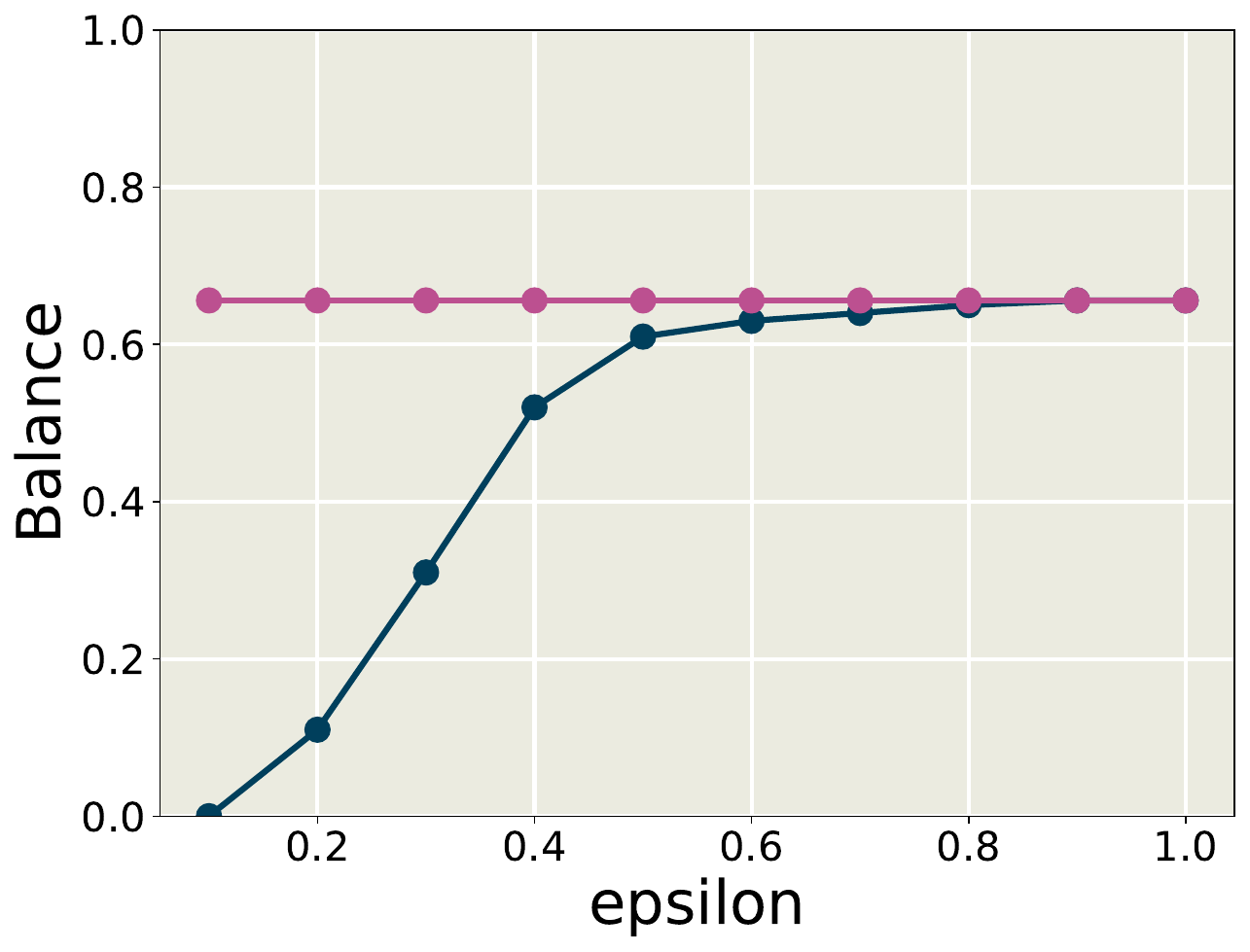}
    \caption{}
    \label{fig:Credit_balance_vs_eps}
  \end{subfigure}
    \begin{subfigure}[b]{0.24\textwidth}
    \includegraphics[width=\textwidth]{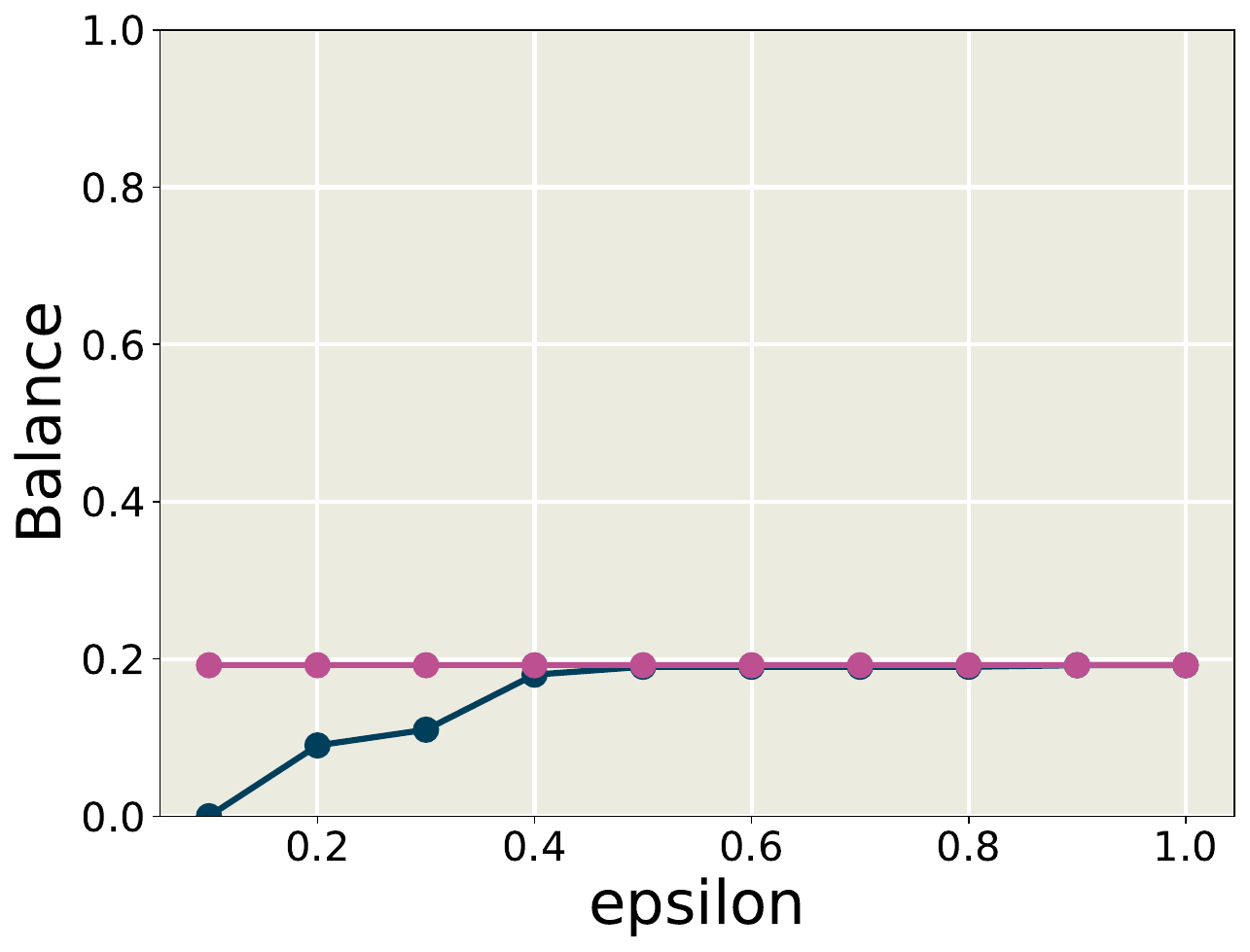}
    \caption{}
    \label{fig:Crime_balance_vs_eps}
  \end{subfigure}
  \begin{subfigure}[b]{0.24\textwidth}
    \includegraphics[width=\textwidth]{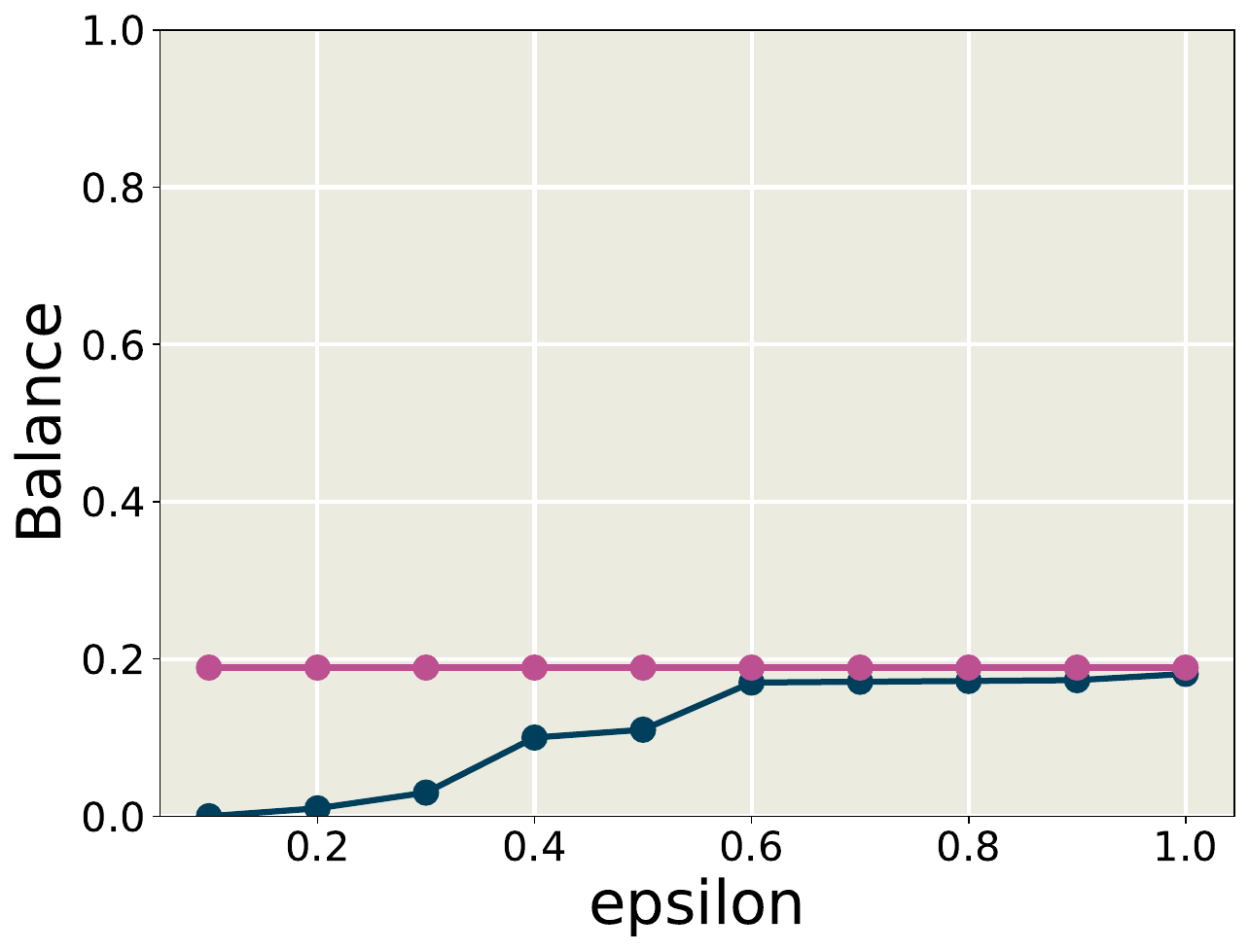}
    \caption{}
    \label{fig:Law_balance_vs_eps}
  \end{subfigure}
  \begin{subfigure}[b]{0.24\textwidth}
    \includegraphics[width=\textwidth]{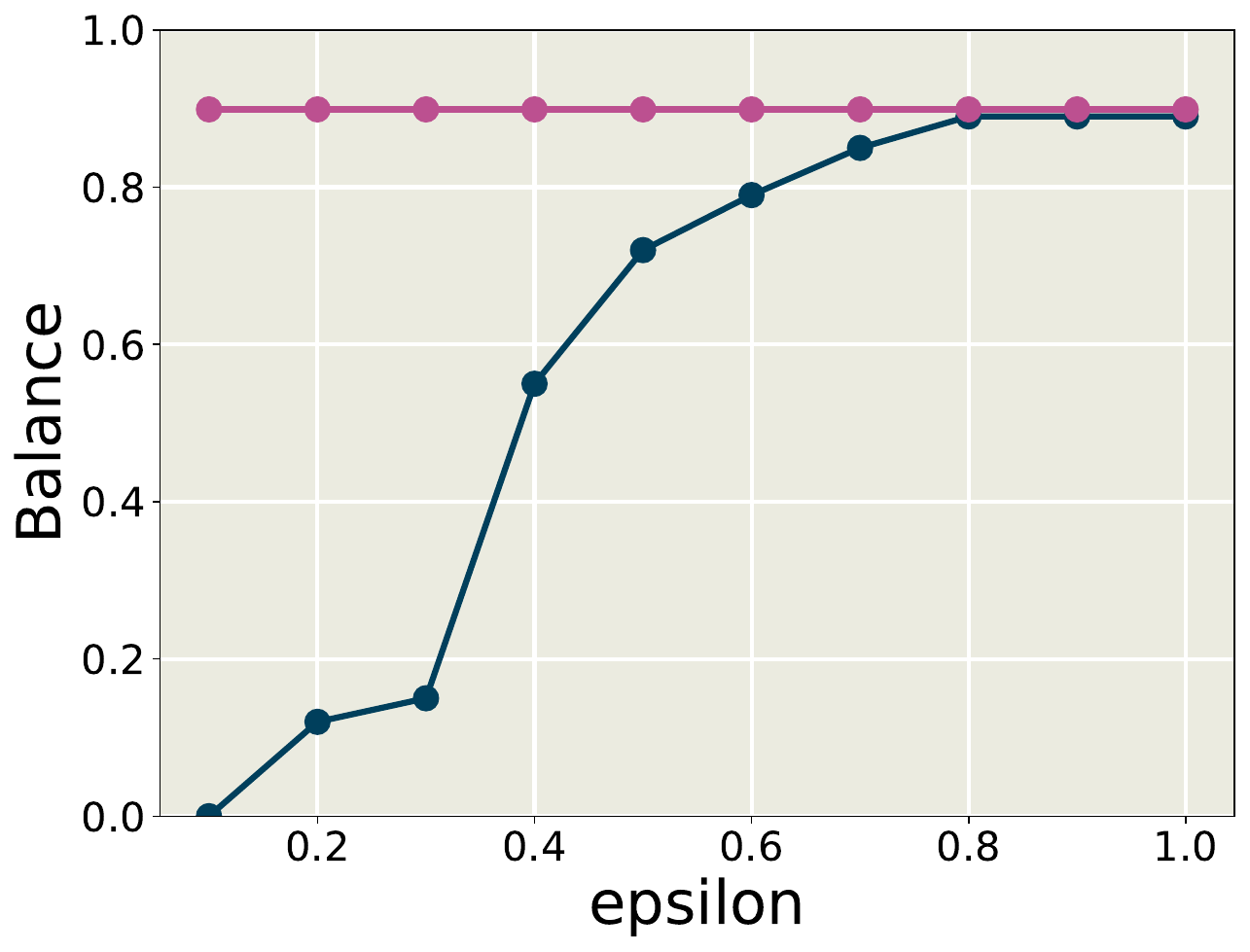}
    \caption{}
    \label{fig:Student_balance_vs_eps}
  \end{subfigure}
 
  \caption{Figures (a) to (h) show the change in Balance with $k$ for fair\_kNN graph construction on the real-world datasets. Figures (i) to (p) show the change in Balance with $\epsilon$ for Fair\_$\epsilon$-neighborhood graph construction on the real-world datasets. In Figures (a) to (h), the x-axis represents the $k$ and the y-axis represents the Balance of the clusters. In Figures (i) to (p), the x-axis represents the value of $\epsilon$ and the y-axis represents the Balance of the clusters. The Figures (a) to (h), each represents a real-world dataset in the given order: Adult, Bank, Catalonia, COMPAS, Credit, Crime, Law, and Student. Similarly, the Figures (i) to (p), each represents a real-world dataset in the given order: Adult, Bank, Catalonia, COMPAS, Credit, Crime, Law, and Student. The pink line represents the maximum achievable Balance for each dataset. The blue line represents the change in Balance when performing the graph construction with fairness constraints.}
  \label{fig:Real_balance_vs_k_eps}
\end{figure*}

\begin{table*}
\centering
\caption{Results on real-world image datasets}
\label{tab:image_dataset_results}
\begin{threeparttable}
\begin{tabular}{|c|cc|cc|cc|} 
\hline
\textbf{Method}                                                     & \multicolumn{2}{c|}{\textbf{MNIST-USPS}} & \multicolumn{2}{c|}{\textbf{Reverse MNIST}} & \multicolumn{2}{c|}{\textbf{MTFL}}  \\ 
\hline
                                                                    & \textbf{Balance} & \textbf{Acc}          & \textbf{Balance} & \textbf{Acc}             & \textbf{Balance} & \textbf{Acc}     \\ 
\hline
\textbf{kNN-SC }                                                    & 0.406            & 0.528                 & 0.219            & 0.756                    & 0.312            & 0.689            \\ 
\hline
\textbf{$\epsilon$-SC}                                                 & 0.712            & 0.618                 & 0.021            & 0.781                    & 0.185            & 0.712            \\ 
\hline
\textbf{RGC}                                                        & 0.000             & 0.412                 & 0.000            & 0.640                     & 0.367            & 0.734            \\ 
\hline
\textbf{SGL}                                                        & 0.000             & 0.435                 & 0.438            & 0.777                    & 0.389            & 0.745            \\ 
\hline
\textbf{CDC}                                                        & 0.611            & 0.593                 & 0.438            & 0.728                    & 0.356            & 0.698            \\ 
\hline
\rowcolor[rgb]{0.871,0.867,0.855} \textbf{SFC}                      & \textbf{1.000}     & 0.295                 & \textbf{1.000}     & 0.442                    & \textbf{0.982}   & 0.667            \\ 
\hline
\rowcolor[rgb]{0.871,0.867,0.855} \textbf{kNN-FSC }                 & 0.000            & 0.012                 & 0.000            & 0.042                    & 0.000            & 0.053            \\ 
\hline
\rowcolor[rgb]{0.871,0.867,0.855} \textbf{$\epsilon$-FSC}              & 0.000            & 0.011                 & 0.000            & 0.01                     & 0.000            & 0.011            \\ 
\hline
\rowcolor[rgb]{0.871,0.867,0.855} \textbf{fair\_kNN-SC (Ours)}      & \textbf{1.000}     & \uline{0.617}         & \uline{0.978}    & \uline{0.751}            & \uline{0.967}    & \uline{0.678}    \\ 
\hline
\rowcolor[rgb]{0.871,0.867,0.855} \textbf{fair\_$\epsilon$-SC (Ours)}  & \uline{0.991}    & \textbf{0.639}        & 0.949            & \textbf{0.769}           & \uline{0.967}    & \textbf{0.701}   \\ 
\hline
\rowcolor[rgb]{0.871,0.867,0.855} \textbf{fair\_kNN-FSC (Ours)}     & 0.000             & 0.032                 & 0.969            & 0.025                    & 0.000            & 0.011            \\ 
\hline
\rowcolor[rgb]{0.871,0.867,0.855} \textbf{fair\_$\epsilon$-FSC (Ours)} & 0.000             & 0.032                 & 0.969            & 0.023                    & 0.000            & 0.072            \\
\hline
\end{tabular}
\begin{tablenotes}
  \small
  \item The Balance metric is calculated for the clusters obtained. In addition, the accuracy of the clustering is obtained. The fair methods are given with a grey background. The best and second-best values for fair methods are highlighted in bold and underline, respectively.
\end{tablenotes}
\end{threeparttable}
\end{table*}

\subsection{Results}
\label{sec:results}
In this section, we present the results of our experiments on both synthetic and real-world datasets. We evaluate the performance of our fair graph construction methods in terms of fairness and clustering quality. In addition, we analyze the impact of different parameters on the performance of our methods, including the disparate impact parameter $\alpha$, the number of neighbors $k$, and the value of $\epsilon$.

\subsubsection{Fairness}

In Table \ref{table:synthetic_results}, we present the results of the experiments conducted on Stochastic Block Model (SBM) synthetic dataset. The dataset is generated with fair ground truth clusters, and we evaluate the performance of our method in terms of clustering error and fairness. From the results, we observe that our Fair\_$\epsilon$-neighborhood graph construction method, when paired with spectral clustering, consistently outperforms the other methods in terms of clustering error and fairness. Even when a uniform error of 0.4 to 0.6 is introduced, our method maintains a lower clustering error compared to the baselines. The fair\_kNN graph construction method also shows competitive performance, and has the same fairness as the Fair\_$\epsilon$-neighborhood graph construction method in most cases. The fact that our method of Fair\_$\epsilon$-neighborhood graph construction is able to achieve better clustering error and fairness is indicative of its ability to uncover fair clusters in the data and construct a graph that densely connects those fair clusters. This is due to the fact that the node addition process in our method is designed to ensure that nodes from different sensitive groups are connected, but keeping the local structure of the graph intact. This allows our method to create a fair graph that is also effective for clustering.

In Table \ref{table:real_world_results}, we present the results of our experiments on real-world tabular datasets. Our proposed methods demonstrate superior performance across multiple fairness and clustering quality metrics. Fair\_$\epsilon$-SC achieves the best Balance scores on several datasets, including Credit (0.656), Crime (0.189), and Student (0.871), while simultaneously maintaining competitive performance on clustering quality metrics. The fair\_kNN-FSC method shows particularly strong performance for Balance on the Adult, Bank, Catalonia, and Compas datasets. E.g., the Balance score for fair\_kNN-SC on the Adult dataset is 0.476, which approaches the target Balance of 0.479. 

Notably, our methods consistently outperform traditional fair clustering approaches. For instance, on the Adult dataset, our Fair\_$\epsilon$-SC achieves a Balance of 0.471 compared to the baseline $\epsilon$-FSC's 0.333. Similarly, on the Bank dataset, fair\_kNN-SC achieves a Balance of 0.449 compared to SFC's 0.442, knn-FSC's 0.438 and $\epsilon$-FSC's 0.301. These results highlight the effectiveness of our fair graph construction methods in achieving fairness consistently across different datasets.
This pattern is followed across most datasets, where our fair graph construction methods outperform the baseline methods in terms of Balance scores. 
On the Crime dataset, our fair\_kNN-SC achieves the highest Balance score of 0.189, significantly outperforming baseline methods like kNN-SC (0.023) and $\epsilon$-SC (0.100). For the Student dataset, fair\_kNN-SC achieves an exceptional Balance score of 0.871, very close to the target Balance of 0.899.

\subsubsection{Clustering Quality}

In Table \ref{table:synthetic_results}, we present the clustering error for the synthetic dataset. As explained above, the dataset is created with fair ground truth clusters. So, deviations from the ground truth clustering also impact the fairness of the clustering. We notice that as the ground truth clusters are fair, our method is performing significantly better in reducing the clustering error. This is indicative of the fact that our method uncovers the fair clusters in the data, constructs a graph, and densely connects those fair clusters.

The real-world tabular dataset results in Table \ref{table:real_world_results} demonstrate that our methods achieve superior clustering quality while maintaining fairness. Fair $\epsilon$-SC consistently delivers the best Silhouette Scores across all datasets: Adult (0.421), Bank (0.433), Catalonia (0.398), Compas (0.367), Credit (0.463), Crime (0.483), Law (0.428), and Student (0.485). This consistent performance indicates that our fairness-aware graph construction does not compromise the intrinsic clustering structure of the data.

The NCut metric results further validate our approach's effectiveness. Fair $\epsilon$-SC achieves the best NCut scores on Adult (0.743), Bank (0.617), Catalonia (0.467), Compas (0.489), Credit (0.442), Crime (0.451), and Student (0.294). These results demonstrate that our method successfully identifies meaningful cluster boundaries while ensuring fair representation across sensitive groups. Comparing with baseline methods, our approaches show substantial improvements. For example, on the Adult dataset, Fair\_$\epsilon$-SC achieves a Silhouette Score of 0.421 compared to traditional $\epsilon$-SC's 0.44, while dramatically improving Balance from 0.092 to 0.471. This demonstrates that our method achieves better fairness without sacrificing clustering quality.

The image dataset results in Table \ref{tab:image_dataset_results} further corroborate our findings. On the MNIST-USPS dataset, Fair\_$\epsilon$-SC achieves the highest accuracy of 0.639 while maintaining excellent Balance (0.991). Similarly, on the Reverse MNIST dataset, it achieves 0.769 accuracy with 0.949 Balance, and on MTFL dataset, it reaches 0.701 accuracy with 0.967 Balance. These results demonstrate the generalizability of our approach across different data modalities and confirm that fair clustering can be achieved without significant accuracy degradation

\subsubsection{Effect of parameters on Fairness}

As discussed previously, the parameter $\alpha$ serves as a fairness control mechanism in our graph construction approach. Figure \ref{fig:DI_Balance} illustrates how varying $\alpha$ values impact cluster Balance across both proposed methods. The results demonstrate a clear positive correlation: higher $\alpha$ values consistently lead to improved cluster Balance for both fair\_kNN and fair\_$\epsilon$-neighborhood graph construction methods. This relationship can be attributed to the underlying mechanism of our approach. As the value of $\alpha$ increases, the algorithm prioritizes establishing connections between nodes from different sensitive groups more aggressively. This enhanced cross-group connectivity results in the formation of densely connected components that inherently contain diverse representation from multiple sensitive groups. The empirical evidence presented in Figure \ref{fig:DI_Balance} validates this theoretical understanding, showing a monotonic increase in cluster Balance as $\alpha$ values rise.

One takeaway from this analysis is that keeping the value of $\alpha$ to be 0.8 is a good choice for achieving fair clustering. Thus, this practically means that the burden of choosing the value of $\alpha$ is not there as it can be set to 0.8 in most cases. This is a significant advantage of our approach, as it simplifies the parameter tuning process for practitioners.

Other parameters, such as the number of neighbors $k$ in the kNN graph construction and the neighborhood radius $\epsilon$ in the $\epsilon$-neighborhood graph construction, also play a crucial role in determining the clustering quality and fairness. However, these parameters' values are typically set based on the domain knowledge or empirical validation rather than being directly tied to fairness control parameter like $\alpha$. Even then, our experiments show that the choice of the values of these parameters does not significantly affect the fairness of the clustering results, as long as those values are chosen appropriately for the specific dataset and clustering task. In Figure \ref{fig:Real_balance_vs_k_eps}, we observe that the changes in the values of $k$ and $\epsilon$ does not lead to significant changes in the Balance scores across different datasets. From Figures \ref{fig:Adult_balance_vs_k} to \ref{fig:Student_balance_vs_k}, we see that the Balance scores remain relatively stable across different values of $k$. In the case of $\epsilon$, Figures \ref{fig:Adult_balance_vs_eps} to \ref{fig:Student_balance_vs_eps} show that for very small values of $\epsilon$, the Balance scores are less, but as the value of $\epsilon$ increases, the Balance scores stabilize. This indicates that our approach is robust to variations in these parameters, further enhancing its practical applicability.

\section{Conclusion}

In this research, we introduced a novel approach for fair clustering that leverages graph construction methods to ensure fair representation across sensitive groups. Our method constructs fair kNN and fair $\epsilon$-neighborhood graphs, which are then used for spectral clustering. The fairness of the clustering is achieved by ensuring that nodes from different sensitive groups are connected in the graph, while preserving the local structure of the data. We proposed two pre-processing methods: fair kNN and fair $\epsilon$-neighborhood graph construction, which can be applied to any graph clustering algorithm. These methods ensure that the resulting clusters are fair with respect to sensitive attributes, while also maintaining high clustering quality. We demonstrated the effectiveness of our proposed pre-processing methods on synthetic and real-world datasets, showing that they achieve superior fairness and clustering quality compared to existing methods. Our approach is particularly advantageous as it does not require additional fairness constraints during the clustering process, making it more efficient and easier to implement.

 

\bibliographystyle{IEEEtran}
\bibliography{sn-bibliography.bib}

\newpage

\end{document}